\documentclass{article}
\usepackage[accepted]{icml2024}

\newcommand{\Def}[0]{\mathrel{\mathop:}=}

\usepackage{pifont}
\newcommand{\cmark}{\textcolor{green}{\ding{51}}}
\newcommand{\xmark}{\textcolor{red}{\ding{55}}}

\usepackage{color, colortbl}
\definecolor{Gray}{gray}{0.93}
\definecolor{Orange}{rgb}{1,0.5,0}
\definecolor{DGray}{gray}{0.83}
\definecolor{LightCyan}{rgb}{0.88,1,1}

\usepackage[T1]{fontenc}

\usepackage{microtype}
\usepackage{graphicx}
\usepackage{subfigure}
\usepackage{booktabs}
\usepackage{hyperref}

\usepackage{wrapfig}
\usepackage{pifont}
\usepackage{color, colortbl}

\usepackage{blindtext}
\usepackage{lipsum}

\usepackage{multirow}
\usepackage{graphicx}
\usepackage{listings}

\usepackage{bbm}

\usepackage [english]{babel}
\usepackage [autostyle, english = american]{csquotes}
\usepackage{amsmath}
\usepackage{amssymb}
\usepackage{mathtools}
\usepackage{amsthm}

\usepackage[capitalize,noabbrev]{cleveref}

\theoremstyle{plain}

\theoremstyle{definition}

\theoremstyle{remark}

\usepackage[textsize=tiny]{todonotes}

\usepackage{pifont}
\usepackage{url}
\usepackage[most]{tcolorbox}
\usepackage{lipsum}
\usepackage{wrapfig}
\usepackage{booktabs}
\usepackage{multirow,mathtools } 

\usepackage{adjustbox}
\MakeOuterQuote{"}

\usepackage{amsmath,amsfonts,bm}

\def\eqref#1{(\ref{#1})}

\def\1{\bm{1}}

\def\rvepsilon{{\mathbf{\epsilon}}}

\def\rva{{\mathbf{a}}}

\def\rvg{{\mathbf{g}}}

\def\rvs{{\mathbf{s}}}

\def\rvv{{\mathbf{v}}}

\def\rvx{{\mathbf{x}}}
\def\rvy{{\mathbf{y}}}
\def\rvz{{\mathbf{z}}}

\DeclareMathAlphabet{\mathsfit}{\encodingdefault}{\sfdefault}{m}{sl}
\SetMathAlphabet{\mathsfit}{bold}{\encodingdefault}{\sfdefault}{bx}{n}

\DeclareMathOperator*{\argmin}{arg\,min}

\DeclareMathOperator{\sign}{sign}

\newcommand{\btau}{\boldsymbol\tau}

\usepackage{url}
\usepackage{array}
\usepackage{microtype}
\usepackage{graphicx}
\usepackage{xcolor}
\usepackage{subfigure}
\usepackage{amsmath}
\usepackage{bm}
\usepackage{dsfont}
\usepackage{mathtools}
\usepackage{nccmath}
\usepackage{amssymb}
\usepackage{multirow}
\usepackage{booktabs}
\usepackage{varwidth}
\usepackage{pifont}
\usepackage{makecell}
\usepackage{wrapfig}
\usepackage{varwidth}
\usepackage{pifont}
\usepackage{makecell}
\usepackage{enumitem}
\usepackage{adjustbox}
\usepackage{graphicx,calc}
\usepackage{amsfonts,amsthm,bm}
\usepackage[flushleft]{threeparttable}
\usepackage{pifont}
\usepackage{ulem}

\PassOptionsToPackage{hyphens}{url}\usepackage{hyperref}

\definecolor{ceruleanblue}{rgb}{0.16, 0.32, 0.75}

\hypersetup{
 colorlinks=true,
 citecolor=ceruleanblue,
 linkcolor=ceruleanblue,
 urlcolor=black}

\newcommand{\revision}[1]{\textcolor{black}{#1}}

\NewEnviron{elaboration}{
\par
\begin{tikzpicture}
\node[rectangle,minimum width=0.44\textwidth] (m) {\begin{minipage}{0.44\textwidth}\BODY\end{minipage}};
\draw[thick] (m.south west) rectangle (m.north east);
\end{tikzpicture}
}

\icmltitlerunning{Revisiting Zeroth-Order Optimization for Memory-Efficient LLM Fine-Tuning: A Benchmark}

\begin{document}

\twocolumn[
\icmltitle{Revisiting Zeroth-Order Optimization for \\Memory-Efficient LLM Fine-Tuning: A Benchmark}



\icmlsetsymbol{equal}{*}

\begin{icmlauthorlist}
\icmlauthor{Yihua Zhang}{equal,msu}
\icmlauthor{Pingzhi Li}{equal,unc}
\icmlauthor{Junyuan Hong}{equal,ut}
\icmlauthor{Jiaxiang Li}{equal,umn}
\icmlauthor{Yimeng Zhang}{msu}
\icmlauthor{Wenqing Zheng}{ut}
\icmlauthor{Pin-Yu Chen}{ibm}
\icmlauthor{Jason D. Lee}{princeton}
\icmlauthor{Wotao Yin}{alibaba}
\icmlauthor{Mingyi Hong}{umn}
\icmlauthor{Zhangyang Wang}{ut}
\icmlauthor{Sijia Liu}{msu,ibm}
\icmlauthor{Tianlong Chen}{unc,mit,harvard}
\end{icmlauthorlist}

\icmlaffiliation{msu}{Michigan State University}
\icmlaffiliation{unc}{The University of North Carolina at Chapel Hill}
\icmlaffiliation{ut}{UT Austin}
\icmlaffiliation{umn}{University of Minnesota Twin Cities}
\icmlaffiliation{princeton}{Princeton University}
\icmlaffiliation{ibm}{IBM Research}
\icmlaffiliation{alibaba}{DAMO Academy, Alibaba Group US}
\icmlaffiliation{mit}{MIT}
\icmlaffiliation{harvard}{Harvard University}

\icmlcorrespondingauthor{Sijia Liu}{liusiji5@msu.edu}
\icmlcorrespondingauthor{Tianlong Chen}{tianlong@cs.unc.edu}

\icmlkeywords{Machine Learning, ICML}

\vskip 0.3in
]



\printAffiliationsAndNotice{\icmlEqualContribution} 

\begin{abstract}
In the evolving landscape of natural language processing (NLP), fine-tuning pre-trained Large Language Models (LLMs) with first-order (FO) optimizers like SGD and Adam has become standard. Yet, as LLMs grow {in size}, the substantial memory overhead from back-propagation (BP) for FO gradient computation presents a significant challenge. Addressing this issue is crucial, especially for applications like on-device training where memory efficiency is paramount. This paper proposes a shift towards BP-free, zeroth-order (ZO) optimization as a solution for reducing memory costs during LLM fine-tuning, building on the initial concept introduced by \citet{malladi2024finetuning}. Unlike traditional ZO-SGD methods, our work expands the exploration to a wider array of ZO optimization techniques, through a comprehensive, first-of-its-kind benchmarking study across five LLM families (Roberta, OPT, LLaMA, Vicuna, Mistral), three task complexities, and five fine-tuning schemes. Our study unveils previously overlooked optimization principles, highlighting the importance of task alignment, the role of the forward gradient method, and the balance between algorithm complexity and fine-tuning performance. We further introduce novel enhancements to ZO optimization, including block-wise descent, hybrid training, and gradient sparsity. Our study offers a promising direction for achieving further memory-efficient LLM fine-tuning. \revision{Codes to reproduce all our experiments are at \url{https://github.com/ZO-Bench/ZO-LLM}.}

\end{abstract}

\section{Introduction}
\label{sec: intro}

Fine-tuning pre-trained large language models~(LLMs) has become the \textit{de-facto} standard in the current paradigms of natural language processing~(NLP)~\cite{raffel2023exploring,sanh2022multitask}. First-order (FO) optimizers, \textit{e.g.}, SGD \cite{amari1993backpropagation}  and  Adam~\cite{kingma2014adam}, have been the predominant choices for LLM fine-tuning. However, as LLMs continue to scale, they encounter significant memory overhead due to the back-propagation (BP) required for FO gradient computation. For example, computing the gradient of the LLM \texttt{OPT-13B} requires $12\times$ more memory cost than the model inference. 
This leads to the challenge of achieving \textit{memory-efficient fine-tuning} in LLMs. Advancements in addressing this challenge could also facilitate technological breakthroughs in related areas, such as on-device training, where memory efficiency is in high demand~\cite{han2015deep,zhu2023ondevice}.

To enhance memory efficiency, an emerging solution is to replace a BP-required FO optimization method with a \textit{BP-free} optimizer during LLM fine-tuning. This was initially proposed by \citet{malladi2024finetuning}, where the FO gradient is approximated using a finite difference of function values.
Despite its new application to LLM fine-tuning, the underlying optimization principle used in \citet{malladi2024finetuning} is commonly known as \textit{zeroth-order (ZO) optimization}, and the function value-based gradient estimate is referred to as the ZO gradient estimate \citep{flaxman2005online,nesterov2017random,duchi2015optimal,ghadimi2013stochastic, liu2020primer}.
 \citet{malladi2024finetuning} employed the classical ZO stochastic gradient descent (ZO-SGD) algorithm \cite{ghadimi2013stochastic}, termed MeZO, to fine-tune the pre-trained LLMs and leveraged the BP-free characteristics of ZO optimization to reduce memory costs.
However, from the perspective of ZO optimization, in addition to ZO-SGD, many other ZO optimization methods have not yet been explored in the context of LLM fine-tuning.
Thus, it remains elusive whether there are potential improvements in  \textit{accuracy} and/or \textit{efficiency} that can be achieved through a benchmarking study of ZO optimization for LLM fine-tuning.
This yields the primary question to be explored:

\begin{center}
\textit{(Q) Can we establish a benchmark for ZO optimization in LLM fine-tuning, explore the overlooked optimization principles, and advance the current state of the art?} 
\vspace*{-3mm}
\end{center}

To address (Q), our work introduces several key innovations compared to the most relevant work \cite{malladi2024finetuning}. We explore a broader range of ZO optimization methods beyond ZO-SGD and examine various task and model types, as well as evaluation metrics. We conduct a detailed comparative analysis of different ZO optimization methods, shedding light on the often-overlooked forward gradient method \cite{ren2022scaling} and other ZO optimization techniques in LLM fine-tuning. This benchmarking study helps reveal the pros and cons of these methods in accuracy and efficiency. Extended from the gained insights, we propose to further improve ZO optimization-based LLM fine-tuning using techniques of block-wise descent, hybrid ZO and FO training, and gradient sparsity. In summary, our key  \textbf{contributions} are listed below.

$\bullet$  We create the first benchmark for ZO optimization in the context of LLM fine-tuning. This benchmark includes our investigations into $6$ BP-free or ZO optimization methods, $5$ LLM families, $3$ tasks of varying complexities, and $5$ fine-tuning schemes, covering both full-parameter and parameter-efficient fine-tuning (PEFT) approaches.

$\bullet$ 
Assisted by our benchmark, we reveal a range of previously overlooked optimization principles and insights for LLM fine-tuning with ZO optimization. These include the significance of aligning tasks to enhance ZO optimization, the role of forward gradient as an LLM fine-tuning baseline, and the trade-offs between algorithm complexity, fine-tuning accuracy, query and memory efficiency.

$\bullet$ In addition to a holistic assessment of existing ZO optimization methods for LLM fine-tuning, we introduce novel enhancements to ZO optimization, including block-wise ZO optimization, hybrid ZO and FO fine-tuning, and sparsity-induced ZO optimization. These proposed techniques aim to improve the accuracy of ZO LLM fine-tuning while maintaining memory efficiency.

\section{Related Work}
\label{sec: related_work}

\textbf{Parameter-efficient fine-tuning (PEFT).} 
Early efforts~\cite{houlsby2019parameter, lin2020exploring} involved inserting trainable adapters, which are compact feed-forward networks, between the layers of the pre-trained model. More recently, various PEFT strategies have been proposed. For instance, \textit{Adapter}-based methods~\cite{houlsby2019parameter, chen2022adaptformer, luo2023towards,karimi2021compacter,pfeiffer2020adapterhub} insert a few tunable yet highly compact modules into pre-trained models. \textit{LoRA}~\cite{hu2021lora} employs trainable low-rank weight perturbations to the pre-trained model, effectively reducing the required number of fine-tuning parameters.
\textit{Prompt-based learning} \cite{gao2020making, hu2021knowledgeable, tan2021msp} has demonstrated effectiveness in various NLP tasks.
Additionally, methods like \textit{prompt tuning}~\cite{lester2021power} and \textit{prefix tuning}~\cite{li2021prefixtuning} incorporate learnable continuous embeddings into the model's hidden states to condition the frozen model for specific downstream tasks. The  following work \cite{liu2021p} demonstrates its applicability on various model scales. While these state-of-the-art PEFT techniques have significantly reduced the number of parameters required for fine-tuning, they still incur memory costs associated with caching numerous activations due to the use of back-propagation (BP)~\citep{malladi2024finetuning}.

\textbf{Zeroth-order optimization.}
Zeroth-order (ZO) optimization is a technique that uses finite differences to estimate gradients. Such algorithms utilize function value oracle only, yet share a similar algorithm structure with first-order (FO) gradient-based counterpart methods. ZO optimization usually enjoys provable (dimension-dependent) convergence guarantees, as discussed in various works \citep{flaxman2005online,nesterov2017random,duchi2015optimal,ghadimi2013stochastic}. These methods have drawn considerable attention due to their effectiveness in a wide range of modern machine learning (ML) challenges \citep{liu2020primer}, including the adversarial attack and defense \citep{chen2017zoo, tu2019autozoom, ye2018hessian, ilyas2018black, zhang2022robustify, verma2023certified, zhao2019design, hogan2018universal, shu2022zeroth}, model-agnostic contrastive explanations \citep{dhurandhar2019model}, enhancing transfer learning through visual prompting \citep{tsai2020transfer}, computational graph unrolling \citep{vicol2023low}, and optimizing automated ML processes \citep{gu2021optimizing,wang2022zarts}. Beyond standard ML, it finds application in policy search in reinforcement learning \citep{vemula2019contrasting}, network resource management \citep{liu2018zeroth}, ML-based optimization of scientific workflows \citep{hoffman2022optimizing,tsaknakis2022zeroth,chen2023deepzero}, and on-chip learning enhancements \citep{gu2021efficient}.

Despite its wide range of use cases, the application of ZO optimization in ML has primarily been restricted to small model scales. This limitation is attributed to the high variance and slow convergence associated with ZO optimization, which are exacerbated by model dimensions.
To scale up ZO optimization, several acceleration techniques have been proposed. These include the integration of historical data to refine ZO gradient estimators \citep{meier2019improving,cheng2021convergence}, leveraging gradient structural information \citep{singhal2023guess} or sparsity to diminish the dependency of ZO optimization on problem size \citep{wang2017stochastic,cai2022zeroth,cai2021zeroth,balasubramanian2018zeroth,ohta2020sparse,gu2021efficient,chen2023deepzero},  the reuse of intermediate features \citep{chen2023deepzero} and random perturbation vectors \cite{malladi2024finetuning} in the optimization process. These advancements suggest a growing potential for the application of ZO optimization in more complex and large-scale ML problems.

\textbf{BP-free training for large models.}
Training large models, especially LLMs, is memory-consuming due to the involved large computation graphs for BP~\citep{ren2021zerooffload,pmlr-v202-kim23l}. Thus, BP-free methods have recently become a focus in the deep learning (DL) community. Forward gradient learning~\cite{baydin2022gradients,ren2022scaling,silver2021learning,belouze2022optimization},  built upon the forward-mode automatic differentiation (AD), provides an alternative to ZO optimization for BP-free training.  
Unlike ZO optimization, it relies on the forward-mode AD to calculate a forward (directional) gradient. 
However, one main limitation of the forward gradient is its requirement of full access to the AD software and the deep model, making it less memory-efficient than ZO optimization and impractical for tackling black-box problems \cite{chen2023deepzero}. The specifications of BP-free methods include greedy layer-wise learning~\cite{nokland2019training},  input-weight alignment \cite{boopathy2022train}, forward-forward algorithm~\cite{hinton2022forward}, synthetic gradients~\cite{jaderberg2017decoupled}, BBT/BBTv2 evolutionary algorithms~\citep{sun2022bbtv2,sun2022black}, gradient guessing using special low dimensional structure of neural networks~\citep{singhal2023guess} and other black-box methods which optimize the prompts~\citep{prasad2023grips,deng2022rlprompt,chai2022cliptuning}. Many of these algorithms are also motivated by seeking DL's biological interpretation. 
 
Applying ZO optimization to fine-tune pre-trained LLMs is particularly intriguing because it combines the advantages of being BP-free and utilizing pre-training. This improves the scalability of ZO optimization to large-scale LLMs while maintaining memory efficiency.
MeZO~\citep{malladi2024finetuning} introduced a ZO-SGD algorithm to fine-tune LLMs with up to 60 billion parameters, very competitive compared to first-order optimization methods and structured fine-tuning approaches like LoRA. They also provided theoretical insights into why ZO methods can be effective for LLMs.  This opens the gateway for efficient BP-free LLM fine-tuning and largely motivates our ZO benchmark study.

\section{Reviewing ZO Optimization and Beyond}
\label{sec:method}
This work's core objective is to benchmark and harness the potential of ZO ({zeroth-order}) optimization in LLM finetuning, eliminating the need for (first-order) back-propagation (BP) during finetuning and thus achieving  {memory efficiency} \cite{malladi2024finetuning}. 
It is worth noting that the ZO optimization technique utilized in \cite{malladi2024finetuning} is primarily the basic version, specifically, ZO stochastic gradient descent (ZO-SGD). There are more advanced ZO optimization methods available, as summarized in \cite{liu2020primer}. Thus, this section is dedicated to reviewing a broader range of ZO optimization approaches and shedding light on the previously overlooked principles for LLM fine-tuning.

\textbf{Basics of ZO optimization.}
ZO optimization serves as a gradient-free alternative to first-order (FO) optimization, approximating FO gradients through function value-based gradient estimates, which we call \textit{ZO gradient estimates}, as discussed in \cite{flaxman2004online,nesterov2017random,ghadimi2013stochastic,duchi2015optimal}. Thus, a ZO optimization method typically mirrors the algorithmic framework of its corresponding FO optimization counterpart. However, it substitutes the FO gradient with the ZO gradient estimate as the descent direction. 

Various techniques exist for performing ZO gradient estimation. In this paper, we focus on the \textit{randomized gradient estimator} (\textbf{RGE}) \cite{nesterov2017random,duchi2015optimal}, a method that relies on the finite difference of function values along a randomly chosen direction vector.
RGE has also been used by \citet{malladi2024finetuning} to achieve memory-efficient fine-tuning for LLMs. Its preference in LLM fine-tuning is attributed to its \textit{query efficiency}, \textit{i.e.}, a low number of function queries.
Given a scalar-valued function $f(\mathbf x)$ where $\mathbf x \in \mathbb R^d$ of dimension $d$, the RGE (referred to as $\hat{\nabla}f(\mathbf x)$) is expressed using central difference: 

\vspace*{-.8em}
{
\small
\begin{align}
    \hat{\nabla}f(\mathbf x) = \frac{1}{q}\sum_{i=1}^q \left [ \frac{f(\mathbf x + \mu \mathbf u_i) - f(\mathbf x - \mu \mathbf u_i)}{2\mu}  \mathbf u_i \right ]
    \tag{RGE}
    \label{eq: RGE}
\end{align}
\vspace*{-1.2em}
}%

where $\mathbf u_i$ is a random direction vector typically drawn from the standard Gaussian distribution $\mathcal N (\mathbf 0, \mathbf I)$, $q$ is the number of function queries, and $\mu > 0$ is a small perturbation stepsize (also known as smoothing parameter). \citet{malladi2024finetuning} employed  \ref{eq: RGE} by setting $q = 1$ and $\mathbf u_i = \mathbf u$. Yet, it is worth noting that the query number $q$ strikes a balance between the ZO gradient estimation variance and the query complexity. It has been shown in \cite{duchi2015optimal,liu2018zeroth} that the variance of RGE is roughly in the order of $O(d/q)$, where $O(\cdot)$ signifies the Big O notation. 

The rationale behind RGE stems from the concept of the \textit{directional derivative} \cite{duchi2015optimal}: As $\mu \to 0$ (letting $q = 1$), the finite difference of function values in \eqref{eq: RGE} approaches $f^\prime (\mathbf x, \mathbf u) \Def \mathbf u^T \nabla f(\mathbf x)$, representing the directional derivative of $f(\mathbf x)$ along the random direction $\mathbf u \sim \mathcal N(\mathbf 0, \mathbf I)$. Subsequently,  \ref{eq: RGE} yields  $\hat{\nabla}f(\mathbf x) \to  f^\prime (\mathbf x, \mathbf u) \mathbf u$ as $\mu \to 0$. Moreover, the directional derivative provides us an unbiased gradient estimator of $\nabla f(\mathbf x)$:

\vspace*{-.8em}
{\small
\begin{align}
  \mathbf{E_{\mathbf u}} [  f^\prime (\mathbf x, \mathbf u) \mathbf u ] 
  = \mathbf{E_{\mathbf u}} [   \mathbf u \mathbf u^T \nabla f(\mathbf x) ]  = \nabla f(\mathbf x).
  \label{eq: unbiased_FG}
\end{align}
\vspace*{-2em}
}%

With the above background, the RGE $\hat{\nabla}f(\mathbf x)$ can be interpreted as an approximation of the FO gradient $\nabla f(\mathbf x)$ using the directional derivative.

\paragraph{Forward gradient: A missing BP-free baseline in LLM fine-tuning.} 
As a byproduct of connecting \ref{eq: RGE} to \eqref{eq: unbiased_FG}, we obtain the directional derivative-based gradient estimate, $\nabla f(\mathbf x) \approx f^\prime (\mathbf x, \mathbf u) \mathbf u$, which is known as the \textit{forward gradient} (\textbf{Forward-Grad}) \cite{baydin2022gradients,ren2022scaling}.
Different from  \ref{eq: RGE} that relies solely on the finite difference of function values, Forward-Grad requires the use of forward mode automatic differentiation (AD) but eliminates the need for backward evaluation in the implementation of deep model fine-tuning or training.
In other words, Forward-Grad is BP-free and can serve as another alternative gradient estimation method that improves the memory efficiency of LLM fine-tuning.
We stress that Forward-Grad is a possibly overlooked BP-free optimizer. Given its unbiasedness as shown in \eqref{eq: unbiased_FG}, it could serve as an upper performance bound for ZO optimization in theory.

\textbf{A focused spectrum of ZO optimization methods.}
Next, we provide a brief overview of the ZO optimization methods to be focused on in this work. Specifically, we will include:
\textbf{ZO-SGD} \cite{ghadimi2013stochastic} that \citet{malladi2024finetuning} has employed for LLM fine-tuning,   ZO-SGD using \underline{sign}-based gradient estimation (\textbf{ZO-SGD-Sign}) \cite{liu2018signsgd}, ZO-SGD with \underline{m}o\underline{m}en\underline{t}um (\textbf{ZO-SGD-MMT}) \cite{malladi2024finetuning}, ZO-SGD with \underline{cons}ervative gradient update (\textbf{ZO-SGD-Cons}), and the ZO variant of the \underline{Adam} optimizer (\textbf{ZO-Adam}) \cite{chen2019zo}.

The aforementioned methods can be unified into the following generic optimization framework in solving $\min_{\mathbf x} f(\mathbf x)$:

\vspace*{-1em}
{\small
\begin{align}
    \mathbf x_{t+1} = \mathbf x_t - \eta_t h( \hat{\nabla}f(\mathbf x_t) ),
    \label{eq: ZO_step}
\end{align}
\vspace*{-2em}
}

where $\mathbf x_t$ denotes the updated solution at the $t$th iteration, $\eta_t >0$ is the learning rate, and $h(\cdot)$ is a certain descent direction post-processing operation.
In \eqref{eq: ZO_step}, we omit the inclusion of the stochastic mini-batch for empirical risk minimization for ease of presentation.
For instance, ZO-SGD can be expressed as \eqref{eq: ZO_step} when $h(\hat{\nabla}f(\mathbf x)) = \hat{\nabla}f(\mathbf x)$. Similarly, ZO-SGD-Sign can be derived if $h(\hat{\nabla}f(\mathbf x)) = \mathrm{sign}(\hat{\nabla}f(\mathbf x))$, where $\mathrm{sign}(\cdot)$ represents element-wise sign operation.
Another example is ZO-SGD-Cons by setting $h(\hat{\nabla}f(\mathbf x)) = \argmin_{\mathbf g \in \{\mathbf 0, -\hat{\nabla}f(\mathbf x), \hat{\nabla}f(\mathbf x) \}} f(\mathbf x_t + \eta_t \mathbf g) $.
We refer readers to Appendix \ref{sec: ZO} for more algorithmic details of \eqref{eq: ZO_step} as applied to the ZO optimization approaches.

\textbf{Our rationale} for selecting the aforementioned ZO optimization approaches for LLM fine-tuning is based on two key considerations: (1) We prioritize ZO optimization methods that require minimal modifications to the existing FO optimizer, ensuring ease of implementation for LLM fine-tuning. (2) We focus on methods with distinct algorithmic characteristics, allowing us to explore a diverse range of optimization strategies for improving LLM performance. Regarding (2), we include ZO-SGD-Sign as it employs 1-bit gradient quantization and represents one of the simplest ZO optimization methods. Additionally, we include ZO-SGD-MMT and ZO-SGD-Cons as they incorporate certain forms of `adaptive learning' into the descent step updates. The former utilizes momentum based on historical gradient information, while the latter allows for the heuristics-based selection of the descent direction. Furthermore, ZO-Adam is one of the most complex ZO optimizers due to its utilization of moving averages and adaptive learning rates.

\textbf{Task alignment in ZO optimization for LLM fine-tuning.} 
Scaling up ZO optimization for deep model training, as discussed in \cite{chen2023deepzero}, is exceedingly challenging due to its high variance, which is dependent on the model size.
Nevertheless, LLM pre-training offers a unique advantage by enabling the fine-tuner to start from a well-optimized pre-trained model state. This graceful model initialization makes ZO optimization potentially scalable to LLM fine-tuning tasks \cite{malladi2024finetuning}.
Even in this pretraining-finetuning paradigm, another crucial factor, which we call `\textbf{task alignment}', still plays a key role in achieving satisfactory ZO fine-tuning performance.
The `task alignment' refers to aligning the fine-tuning task with the format of the pre-training task, given by the next token or sentence prediction. 
For example, \citet{gao2020making, malladi2024finetuning} have transformed downstream text classification tasks into next token prediction tasks by introducing well-crafted input prompts.
These prompts serve as bridges to align the fine-tuning tasks with the pre-training ones, facilitating ZO optimization when initiated from the pre-trained model.

\begin{table}[t]
\centering
\vspace{-3mm}
\caption{Test accuracy (\%) of pretrained Roberta-Large model fine-tuned on SST2 and RTE tasks using ZO and FO optimization methods with ({\cmark}) and without ({\xmark}) text alignment. The evident performance degradation is highlighted \textbf{in bold}.}
\label{tab: text_align_preliminary}
\resizebox{.8\linewidth}{!}{%
\begin{tabular}{c|ccc|ccc}
\toprule[1pt]
\midrule
\multirow{2}{*}{Method} 
& \multicolumn{3}{c|}{SST2} 
& \multicolumn{3}{c}{RTE} \\ \cmidrule{2-4} \cmidrule{5-7}
& {\cmark} & {\xmark} & Difference & {\cmark} & {\xmark}  & Difference \\
\midrule
FO-SGD
& 91.6
& 91.5
& 0.1
& 70.9
& 61.4
& 9.5
\\
\midrule
ZO-SGD
& 89.4
& 79.2
& \textbf{10.2}
& 68.7
& 60.4
& \textbf{8.3}
\\
ZO-Adam
& 89.8
& 79.2
& \textbf{10.6}
& 69.2
& 58.7
& \textbf{10.5}
\\
\midrule
\bottomrule[1pt]
\end{tabular}%
}
\vspace{-2.5em}
\end{table}

As a warm-up experiment, \textbf{Tab.\,\ref{tab: text_align_preliminary}} empirically justifies the importance of task alignment when applying ZO optimization to LLM fine-tuning on the simple binary classification task by comparing scenarios \textit{with} and \textit{without}  the use of pre-defined prompts to achieve task alignment. We fine-tune the {entire} Roberta-Large \cite{liu2019roberta} model on SST2 \cite{socher2013recursive} and RTE \cite{wang2019glue} datasets with two selected ZO methods: ZO-SGD (\textit{i.e.}, MeZO in \cite{malladi2024finetuning}) and ZO-Adam. We compare their performance with that of the FO method (FO-SGD). The task alignment is achieved with the template \texttt{<CLS>SENTENCE. It was [terrible|great].<SEP>} for the SST dataset and another template \texttt{<CLS>SENTENCE1? [Yes|No], SENTENCE2.<SEP>} for RTE. 
As we can see, without prompt-based text alignment, there is a substantial performance drop across ZO fine-tuning methods.  Both ZO-SGD and ZO-Adam yield about $10\%$ and $8\%$  accuracy degradation on SST2 and RTE, respectively. In contrast, FO-SGD suffers less from the absence of task alignment. 
This suggests that the task alignment is particularly beneficial to ZO LLM fine-tuning. 
It is also worth noting that crafting effective prompts for task alignment can be non-trivial, as prompt design is context-dependent and can affect the fine-tuning performance.
In this work, we follow \cite{gao2020making} and \cite{malladi2024finetuning} to align the fine-tuning tasks to the pretrained ones.

\section{LLM Fine-Tuning Benchmarking}
\label{sec: benchmark}

In this section, we delve into the empirical performance of ZO optimization in LLM fine-tuning. Our benchmarking effort includes evaluating accuracy and efficiency, accounting for different downstream task complexities (ranging from simple classification to more complicated reasoning tasks), and considering various language model types.

\subsection{Benchmark Setups}

\textbf{LLM fine-tuning tasks, schemes, and models.}
We begin by introducing the tasks and the fine-tuning schemes. 
We focus on \textit{three tasks}, considering their complexity from low to high, which include   (1) the simplest binary classification task, Stanford Sentiment Treebank v2 (SST2) \cite{socher2013recursive}, (2) the question-answering task, Choice Of Plausible Alternatives (COPA) \cite{roemmele2011choice}, (3) the commonsense reasoning task, WinoGrande \cite{sakaguchi2021winogrande}, and {(4) the multi-sentence reading comprehension (MultiRC) \cite{MultiRC2018} (for efficiency evaluation only).}
For LLM fine-tuning on these tasks,  we explore \textit{four parameter-efficient fine-tuning (PEFT) schemes}: full-tuning (FT) that fine-tunes the entire pre-trained model, low-rank adaptation (LoRA) by imposing low-rank weight perturbations \cite{hu2021lora}, prefix-tuning (Prefix) by appending learnable parameters to token embedding \cite{li2021prefixtuning}, and prompt-tuning (Prompt) \cite{liu2022p} by introducing a series of learnable tokens attached to the input to adapt the fixed model to downstream tasks.  We refer readers to Appx.\,\ref{sec:pre_peft} for details.
Furthermore, we incorporate several representative language models, including  {Roberta-Large} \cite{liu2019roberta},  OPT \cite{zhang2022opt}, LLaMA2 \cite{touvron2023llama}, Vicuna \cite{zheng2023judging}, and Mistral \cite{jiang2023mistral}.

\textbf{Setup and implementation details.}
To train the previously mentioned LLM fine-tuners, we utilize the ZO optimization methods introduced in Sec.\,\ref{sec:method}. These include ZO-SGD (\textit{i.e.} MeZO \cite{malladi2024finetuning}), 
ZO-SGD-Sign, ZO-SGD-MMT, ZO-SGD-Cons, and ZO-Adam. For comparison, we also present the performance of Forward-Grad, which relies on the forward mode auto-differentiation rather than BP. We also provide the performance of two FO optimizers: (FO-)SGD and (FO-)Adam. 
Before applying the aforementioned optimizers to the LLM fine-tuning tasks, we follow \cite{gao2020making, malladi2024finetuning} to align the fine-tuning task format with the token or sentence prediction-based pre-training task, as demonstrated in Tab.\,\ref{tab: text_align_preliminary}.
We run ZO (or BP-free) optimizers and FO optimizers for 20000 and 625 iterations respectively, as outlined in \eqref{eq: ZO_step}. 
Note that ZO optimization takes a longer convergence time as shown in  \cite{liu2020primer}.
When implementing \eqref{eq: RGE}, unless specified otherwise, we set the query budget per gradient estimation to $q = 1$.
We determine the values of other hyperparameters, such as the smoothing parameter and learning rate, through a grid search for each method.
\revision{Unless otherwise stated, following \cite{malladi2024finetuning}, half-precision training (F16) and mixed-precision training (FP16) are adopted by default on ZO and FO methods, respectively, to reduce the memory consumption. For FP16, the model is loaded with full precision while the training is carried out in half-precision, whereas F16 means both the model and training are in float16. More implementation details can be found in Appx.\,\ref{sec: implementation_details}, and their influence on memory efficiency is discussed in Sec.\,\ref{sec: memory_profile}.}

\textbf{Evaluation metrics.} 
We evaluate ZO LLM fine-tuning using two sets of metrics: accuracy and efficiency. Accuracy measures the fine-tuned model's test-time performance in specific tasks, such as test accuracy in classification tasks. Efficiency includes various measurements like memory efficiency (in terms of peak memory usage and GPU cost), query efficiency (\textit{i.e.}, the number of function queries required for ZO optimization ), and run-time efficiency. These metrics collectively provide insights into the resources needed for ZO LLM fine-tuning, helping assess its feasibility and cost-effectiveness in practical scenarios.

\subsection{Experiment Results}

\begin{table}[t]
    \vspace*{-3mm}
    \centering
    \caption{Performance of LLM fine-tuning on  SST2 over pretrained Roberta-Large and OPT/1.3B. Best performance among ZO methods (including Forward-Grad) is highlighted in \textbf{bold}.}
    \label{tab: preliminary_study}
    \resizebox{\linewidth}{!}{
    \begin{tabular}{lcccccccc}
    \toprule[1pt]
    \midrule
    \multirow{2}{*}{SST2} & \multicolumn{4}{c}{Roberta-Large} & \multicolumn{4}{c}{OPT-1.3B} \\
    \cmidrule(lr){2-5} \cmidrule(lr){6-9}
    & FT & LoRA & Prefix & Prompt & FT & LoRA & Prefix & Prompt \\
    \midrule
    FO-SGD      
    & $91.4$ & $91.2$ & $89.6$ & $90.3$ 
    & $91.1$ & $93.6$ & $93.1$ & $92.8$ \\
    \midrule
    Forward-Grad 
    & $\mathbf{90.1}$ & $89.7$ & $89.5$ & $87.3$ 
    & $90.3$ & $90.3$ & $90.0$ & $82.4$ \\
    \midrule
    ZO-SGD
    & $89.4$ & $90.8$ & $90.0$ & $87.6$ 
    & $\mathbf{90.8}$ & $90.1$ & $\mathbf{91.4}$ & $84.4$ \\
    ZO-SGD-MMT   
    & $89.6$ & $90.9$ & $90.1$ & $88.6$ 
    & $85.2$ & $91.3$ & $91.2$ & $\mathbf{86.9}$ \\
    ZO-SGD-Cons  
    & $89.6$ & $\mathbf{91.6}$ & $90.1$ & $88.5$ 
    & $88.3$ & $90.5$ & $81.8$ & $84.7$ \\
    ZO-SGD-Sign  
    & $52.5$ & $90.2$ & $53.6$ & $86.1$ 
    & $87.2$ & $91.5$ & $89.5$ & $72.9$ \\
    ZO-Adam      
    & $89.8$ & $89.5$ & $\mathbf{90.2}$ & $\mathbf{88.8}$ 
    & $84.4$ & $\mathbf{92.3}$ & $\mathbf{91.4}$ & $75.7$ \\
    \midrule
    \bottomrule[1pt]
    \end{tabular}
    }
    \vspace*{-2.5em}
\end{table}

\textbf{ZO fine-tuning on SST2: A pilot study.} In \textbf{Tab.\,\ref{tab: preliminary_study}}, experiments are conducted to compare the performance of different BP-free and BP-based (FO-SGD) methods on one of the simplest LLM downstream task: the binary classification task with SST2 dataset. We investigate two model architectures, the medium-sized Roberta-Large and the larger model OPT-1.3B. Several key results are summarized below. 

\underline{First}, ZO-Adam seems to be the most effective ZO method, achieving the best performance in 4 out of 8 fine-tuning settings. However, as will be shown later, this is achieved at the cost of additional memory consumption. This is not surprising considering that ZO-Adam has the highest algorithmic complexity, as explained in Sec.\,\ref{sec:method}.

\underline{Second}, Forward-Grad is a competitive method compared to the ZO methods, especially in the FT (full-tuning) setting. This indicates that Forward-Grad may be suitable for problems of a larger scale, and make it a compelling baseline of ZO LLM fine-tuning.
Additionally, when the complexity of the fine-tuning scheme decreases (\textit{e.g.}, Prompt), the advantage of Forward-Grad over function value-based ZO methods diminishes.

\underline{Third}, The performance of ZO methods exhibits high variance, as evidenced by the fluctuating relative rankings across different experimental scenarios, although extensive hyper-parameter search efforts have been made.
For example, the effectiveness of ZO-Adam degrades dramatically in the (OPT-1.3B, Prompt) setting.
In addition, the MeZO method (\textit{i.e.}, ZO-SGD) used in \cite{malladi2024finetuning}
does \textit{not} always emerge as the top-performing ZO optimizer for LLM fine-tuning across various settings.
This is not surprising and could be largely attributed to the high variance of \ref{eq: RGE}  \cite{nesterov2017random,duchi2015optimal}.

\underline{Fourth},  ZO-SGD-Cons and ZO-SGD-MMT also demonstrate strong performance as ZO optimizers in LLM fine-tuning. However, ZO-SGD-Sign, the simplest ZO optimization method, tends to be the weakest approach, except in the simplest fine-tuning setting Prompt.
The above observations motivate us to extend our explorations, investigating the effectiveness of ZO methods across a broader spectrum of models and more complex tasks.

\begin{figure}[t]
\vspace*{-.5em}
\centering
\includegraphics[width=\linewidth]{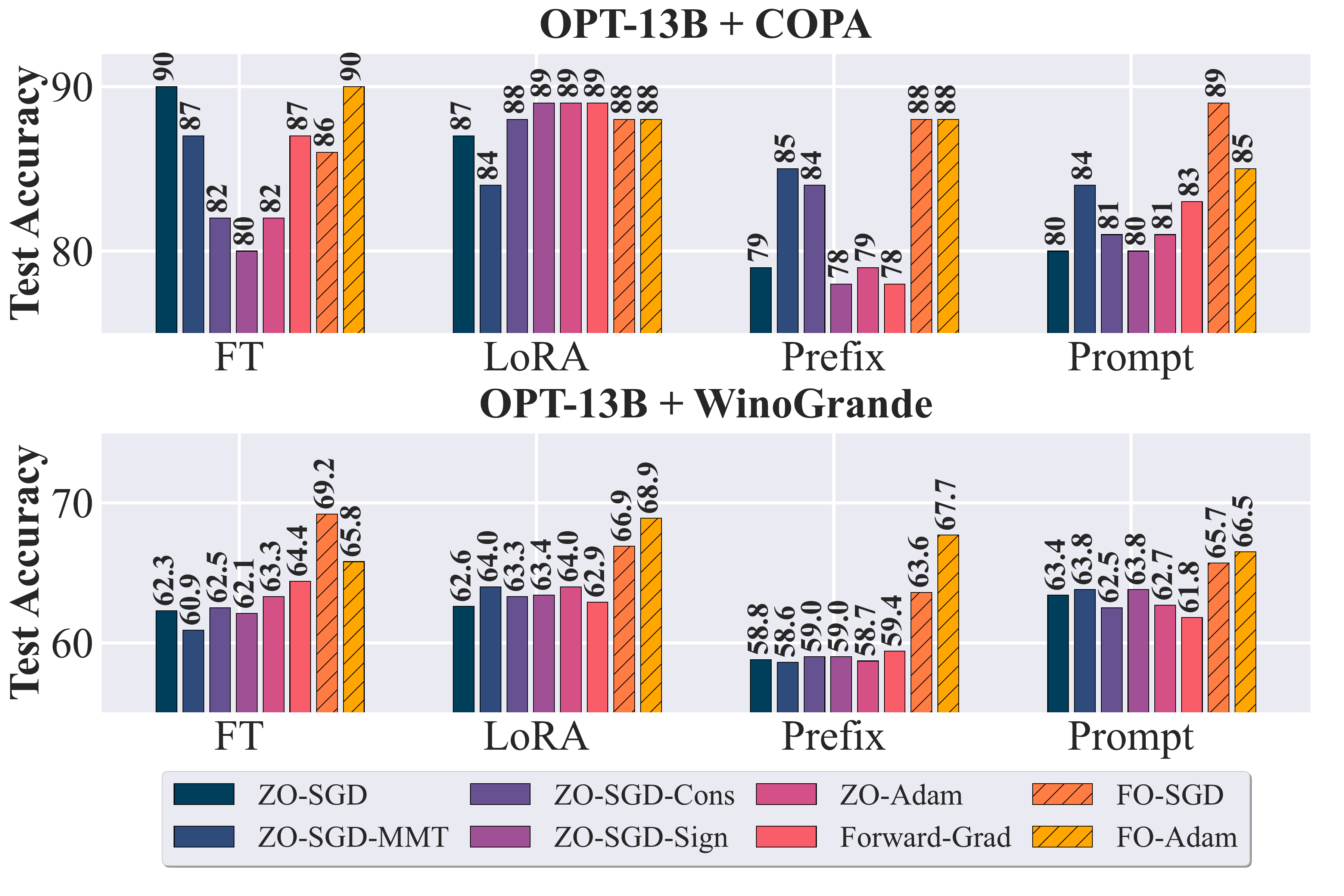}
\vspace*{-2em}
\caption{Results of OPT-13B on the tasks COPA and WinoGrande fine-tuned using ZO/FO optimizers in different PEFT settings.}
\vspace{-2em}
\label{fig: main_result}
\end{figure}

\textbf{ZO fine-tuning on downstream tasks COPA and WinoGrande under OPT-13B.} 
Extended from the experiments on SST2,  \textbf{Fig.\,\ref{fig: main_result}} presents the fine-tuning performance on COPA and WinoGrande dataset using a larger model, OPT-13B. 
We summarize our key observations when the problem scales up and becomes more complicated.

\underline{First}, compared to the previous results, the performance gap among different ZO methods is much enlarged. In the meantime, 
the performance gap between FO and ZO methods is also widened. For example, in the experiments with WinoGrande, the FO methods (FO-SGD and FO-Adam) outperform all the other ZO methods by a large margin. This observation shows the scalability bottleneck intrinsic to ZO methods, when dealing with larger models and/or more complex tasks. 

\underline{Second}, certain ZO methods exhibit exceptional stability across varied conditions: Despite a general trend towards variability, specific ZO methods, \textit{e.g.}, ZO-Adam and ZO-SGD-MMT, demonstrate stability in their performance.  This could be because these algorithms integrate variance-reduced optimization techniques (such as momentum and adaptive learning rate) to ZO optimization and become more adaptable and resilient to the variances of ZO gradient estimation \cite{chen2019zo}.

\underline{Third}, the LoRA tuning is consistently robust when paired with various ZO algorithms. This resilience across different ZO methods suggests that LoRA's mechanism is inherently more adaptable to various ZO optimization strategies, providing a stable and reliable tuning approach in diverse settings. We will peer into the performance of LoRA below.

\begin{table}[t]
\vspace*{-1em}
\centering
\caption{Performance of different LLMs finetuned with LoRA on COPA and WinoGrande  using different ZO/FO methods. Table format is consistent with Tab.\,\ref{tab: preliminary_study}.}
\label{tab: result_model}
\resizebox{.9\linewidth}{!}{%
\begin{tabular}{lcccc}
\toprule[1pt]
\midrule
\multicolumn{1}{c}{} & OPT-13B & LLaMA2-7B & Vicuna-7B & Mistral-7B \\
\midrule
\multicolumn{5}{c}{COPA}                                                        \\
\midrule
FO-SGD               & $88$      & $85$                 & $84$        & $90$         \\
FO-Adam              & $88$      & $84$                 & $81$        & $90$         \\
\midrule
Forward-Grad         & $\mathbf{89}$      & $82$              & $\mathbf{84}$        & $88$         \\
\midrule
ZO-SGD               & $87$      & $\mathbf{86}$             & $83$        & $\mathbf{90}$         \\
ZO-SGD-CONS          & $88$      & $85$             & $83$        & $89$         \\
ZO-Adam              & $\mathbf{89}$      & $83$             & $\mathbf{84}$        & $89$         \\
\midrule
\multicolumn{5}{c}{WinoGrande}                                                        \\
\midrule
FO-SGD               & $66.9$    & $66.9$            & $66.5$      & $76.4$       \\
FO-Adam              & $68.9$    & $69.5$             & $70.0$      & $76.9$       \\
\midrule
Forward-Grad         & $62.9$    & $64.3$             & $\mathbf{65.6}$      & $\mathbf{70.1}$      \\
\midrule
ZO-SGD               & $62.6$    & $64.3$             & $\mathbf{65.6}$      & $68.7$       \\
ZO-SGD-CONS          & $63.3$    & $\mathbf{64.6}$          & $65.3$      & $68.5$       \\
ZO-Adam              & $\mathbf{64.0}$    & $64.4$           & $65.5$      & $69.5$       \\
\midrule
\bottomrule[1pt]
\end{tabular}%
}
\vspace*{-2em}
\end{table}

In \textbf{Tab.\,\ref{tab: result_model}}, we present how different ZO methods perform on (LoRA, COPA) and (LoRA, WinoGrande) across a wide range of LLM families. 
For ease of computation, we focus on a subset of ZO optimization methods, including ZO-SGD, ZO-SGD-CONS, and ZO-Adam. 
As we can see, in some scenarios with the COPA dataset, some BP-free methods exhibit effectiveness comparable to, or even superior to, that of FO methods (FO-SGD and FO-Adam).
For example, Forward-Grad and ZO-Adam outperform the best FO method on model OPT-13B and Vicuna-7B. Conversely, for the more difficult task WinoGrande, a performance gap of $5\%\sim 6\%$ between FO and ZO methods still exists across different models.

\textbf{Efficiency analysis.} 
In \textbf{Tab.\,\ref{tab: efficiency}}, we present a comparison of the efficiency performance of various ZO/FO optimizers when fine-tuning the full {OPT-13B} model on the MultiRC dataset with a batch size of $4$.
We evaluate the efficiency in the following dimensions: memory cost (in GB),  the consumption of GPU resources (number of GPUs), and runtime cost per optimization iteration (in seconds). All the experiments are carried out in the same environment.
\underline{First}, from a memory efficiency standpoint, ZO-SGD, ZO-SGD-Cons, and ZO-SGD-Sign exhibit similar levels of efficiency and only necessitate a \textit{single} GPU (A100) for LLM fine-tuning 
This is not surprising since these ZO methods employ relatively simple optimization steps and primarily rely on the utilization of \ref{eq: RGE}.
\underline{Second}, Forward-Grad seems to be the threshold beyond which the ZO optimization methods lose their memory efficiency advantage over FO methods, as seen with ZO-Adam, for example.
\underline{Third}, compared to FO methods, ZO optimization reduces runtime costs per iteration by $\sim 41.9\%$ for ZO-SGD versus FO-SGD. 

\begin{table}[t]
\vspace*{-1em}
\centering
\caption{The peak memory cost (in GB), the required GPU resources, and the runtime cost (in seconds) of each optimizer when fine-tuning the full {OPT-13B} model on MultiRC with an averaged 400 context length. The order of included optimizers is ranked based on the memory cost. The per-iteration runtime in seconds (s) is averaged over 100 iterations.}
\label{tab: efficiency}
\resizebox{\linewidth}{!}{
\begin{tabular}{l|ccc}
\toprule[1pt]
\midrule
Optimizer    & Memory $\Downarrow$ & Consumed GPUs $\Downarrow$     & Runtime Cost \\ 
\midrule
ZO-SGD       & $\mathbf{64}$ GB   & $\mathbf 1\times$A100   & $\mathbf{11.1}$s    \\
ZO-SGD-Cons  & $\mathbf{64}$ GB   & $\mathbf 1\times$A100   & $27.6$s    \\
ZO-SGD-Sign  & $\mathbf{64}$ GB   & $\mathbf 1\times$A100   & $\mathbf{11.1}$s    \\
ZO-SGD-MMT   & $100$ GB  & $2\times$A100   & $\mathbf{11.1}$s    \\
Forward-Grad & $134$ GB  & $2\times$A100   & $16.7$s    \\
FO-SGD       & $148$ GB  & $2\times$A100   & $19.1$s    \\
ZO-Adam      & $158$ GB  & $2\times$A100   & $\mathbf{11.5}$s    \\
FO-Adam      & $245$ GB  & $4\times$A100   & $19.4$s   \\
\midrule
\bottomrule[1pt]
\end{tabular}
}
\vspace*{-2em}
\end{table}

\begin{figure}[t]
\vspace*{-.5em}
\centering
\includegraphics[width=.8\linewidth]{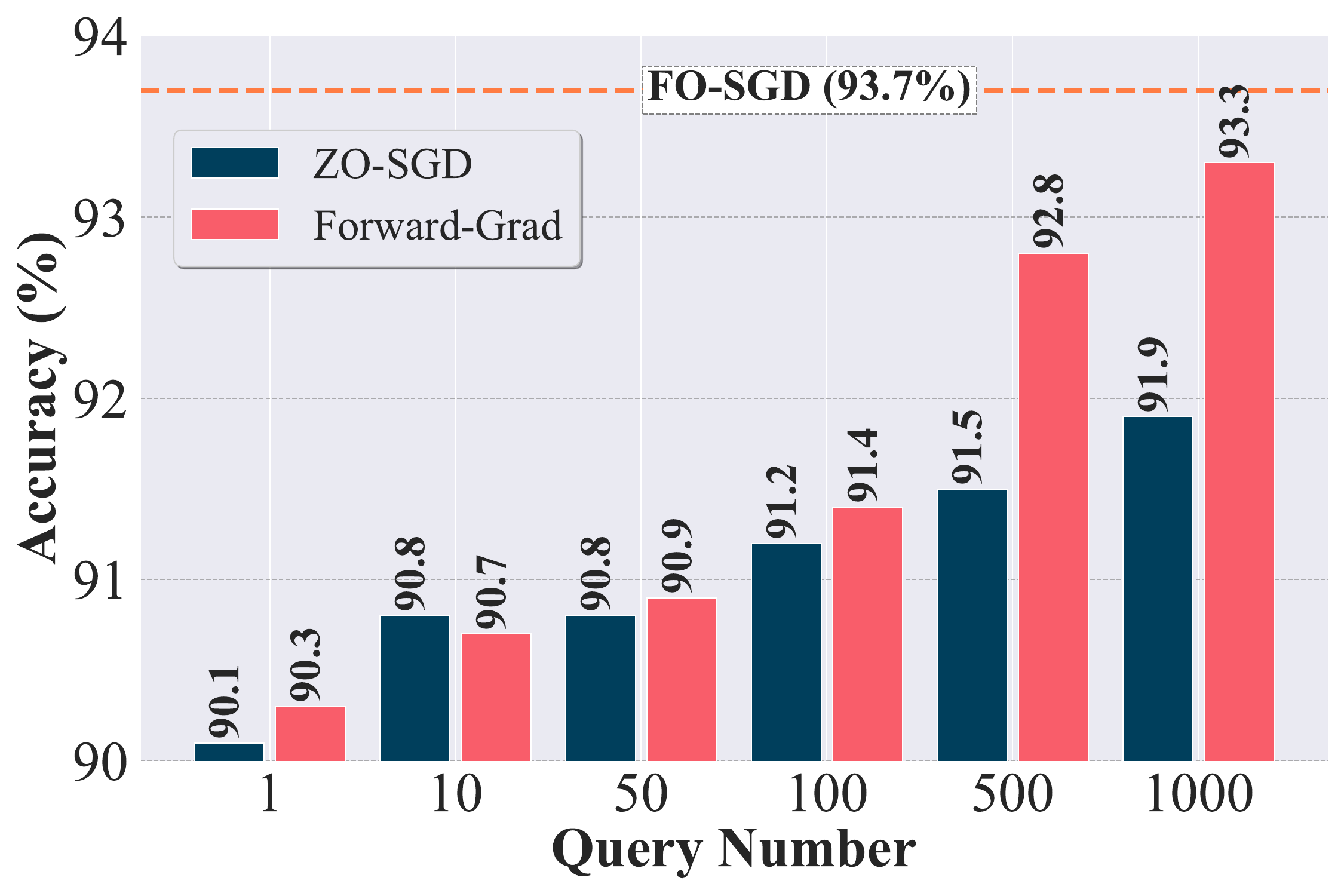}
\vspace{-1em}
\caption{LoRA-based fine-tuning accuracy of OPT-1.3B on SST2 using ZO-SGD and Forward-Grad over different budgets.}
\vspace{-1.5em}
\label{fig: query_number}
\end{figure}

\textbf{Ablation study on query budget.} 
Recall from \eqref{eq: unbiased_FG} that Forward-Grad provides us with an unbiased gradient estimate with respect to the FO gradient, in contrast to the function value-based biased ZO gradient estimate. However, in the above experiments, we have not observed a significant advantage brought by Forward-Grad. We hypothesize that this is due to the smallest query budget $q = 1$ we used, which, despite its query efficiency, may not fully showcase the unbiased benefit. Inspired by the above, 
\textbf{Fig.\,\ref{fig: query_number}} explores the impact of varying query budget ($q$) on the performance of Forward-Grad and ZO-SGD (\textit{i.e.}, MeZO). As we can see, both the accuracies of Forward-Grad and ZO-SGD  are improved with an increased query budget. However, 
the performance rise for Forward-Grad is much more evident. For example, when the query number is larger than $500$, Forward-Grad outperforms ZO-SGD by a large margin $1\%\sim2\%$ and approaches FO-SGD. These observations underscore that the inherent advantages of Forward-Grad become apparent only when a sufficient number of queries are used. The downside of using a higher query budget is a higher computation cost, which scales linearly with $q$.  

\begin{table*}[th]
    \vspace*{-1em}
    \centering
    \caption{Comparison of the instant peak memory consumption of different optimizers when fine-tuning the full model. Peak memory consumption consists of the constant memory for the model (Weight Mem.), the optimizer states (Opt. State Mem.), and dynamic allocation for computing gradients and optimization (Dynamic Mem.). The {cardinalities} $|\rvx|$ and $|\rva|$ correspond to the memory consumption of loading model parameters and saving the intermediate results for post-hoc backward during forward in full precision, with the subscript $l$ representing that of a specific layer $l$. In Dynamic Mem., 
    $\mathbf x$ is also used to denote the memory consumption for temporarily saving the gradients, as they share the same size with the corresponding parameters.}
    \label{tab:total_mem_com}
    \resizebox{.7\linewidth}{!}{
    \begin{tabular}{l|cccc}
    \toprule
    \midrule
    Optimizer & Weight Mem. & Dynamic Mem (Grad.\&Opt.) & Opt. State Mem. \\ \midrule
    \multicolumn{4}{c}{Optimizer in Full Precision}\\ 
    \midrule
    FO-SGD     & $|\rvx|$ & $\sum_l \max\{|\rva_l|, |\rvx_l|\}$ & 0 \\
    FO-Adam w/o fast foreach  & $|\rvx|$ & $\sum_l \max\{|\rva_l|, |\rvx_l|\}$ & $2|\rvx|$ \\
    FO-Adam  & $|\rvx|$ & $\sum_l \max\{|\rva_l|, |\rvx_l|\}$ & $3|\rvx|$ \\
    Forward-Grad & $|\rvx|$ & $|\rvx| + \max_l |\rva_l|$ & $0$  \\
    Vanilla ZO-SGD     & $|\rvx|$ & $|\rvx|$ & 0 \\
    ZO-SGD     & $|\rvx|$ & $\max_l |\rvx_l|$ & 0 \\
    ZO-SGD MMT & $|\rvx|$ & $\max_l |\rvx_l|$ & $|\rvx|$ \\
    ZO-Adam & $|\rvx|$ & $\max_l |\rvx_l|$ & $2|\rvx|$ \\
    \midrule
    
    \multicolumn{4}{c}{Optimizer in Mixed Precision Training (FP16)} \\
    \midrule
    FO-SGD (\textbf{FP16})    & $|\rvx|$ & $\max \left\{\frac {1}{2}|\rva|+\frac {1}{2}|\rvx|, \sum_l \max\{\frac{1}{2} |\rva_l|, |\rvx_l|\}\right\}$ & 0 \\
    FO-Adam (\textbf{FP16})    & $|\rvx|$  & $\max \left\{\frac {1}{2}|\rva|+\frac {1}{2}|\rvx|, \sum_l \max\{\frac{1}{2} |\rva_l|, |\rvx_l|\}\right\}$ & $2|\rvx|$ \\ 
    \midrule
    
    \multicolumn{4}{c}{Optimizer with Half Precision Model (F16)} \\ \midrule
    ZO-SGD (\textbf{F16}) & $\frac{1}{2} |\rvx|$ & $\max_l \frac{1}{2} |\rvx_l|$ & 0  \\
    ZO-SGD-MMT (\textbf{F16}) & $\frac{1}{2} |\rvx|$ & $\max_l \frac{1}{2} |\rvx_l|$ & $\frac{1}{2} |\rvx|$  \\
    ZO-Adam (\textbf{F16}) & $\frac{1}{2} |\rvx|$ & $\max_l \frac{1}{2} |\rvx_l|$ & $|\rvx|$  \\
    \midrule
    \bottomrule
    \end{tabular}}
    \vspace*{-1.5em}
\end{table*}

\subsection{\revision{An In-Depth Dissection on Memory Efficiency}}
\label{sec: memory_profile}
In this section, we will provide provide a holistic memory profile for the ZO and FO methods, theoretically and empirically. And we will discuss how memory efficiency will be influenced by the implementation details adopted in this work, including F16 (half-precision model) and FP16 (mixed-precision training).

\textbf{Theoretical analysis.} The theoretical memory efficiency of all the methods with and without the memory-saving tricks are listed in \textbf{Tab.\,\ref{tab:total_mem_com}}. We refer readers to Appx.\,\ref{sec: theoretical_analysis_mem} for a detailed analysis of these results. Several key insights can be summarized. 
\underline{First}, compared to FO optimizers (including Forward-Grad), the memory efficiency of ZO optimizers mainly comes from two perspectives: (1) ZO methods can avoid saving the intermediate results (model states) $\mathbf{a}_l$. As we will see later, this reduction is considerable when the sequence length of the language model is large; (2) ZO methods can estimate the gradient in a layer-wise manner and thus avoid storing the gradients of the entire model. \underline{Second}, the usage of FP16 in FO methods can only reduce the memory consumption of intermediate results (\textit{i.e.}, $|\rva|$ in Tab.\,\ref{tab:total_mem_com}), 
but not that from the model loading or optimization states' storage. In contrast, ZO methods can be easily equipped with F16, reducing memory by $50\%$. \underline{Third}, although Forward-Grad is also a BP-free method, its memory efficiency advantage over other FO optimizers is not remarkable, mainly due to its requirement for storing the intermediate results (\textit{i.e.}, the computational graph). \underline{Last}, although ZO-Adam is memory intensive compared to ZO-SGD, it can be greatly improved by using F16 and achieves a better efficiency than FO-SGD (FP16).

\textbf{Empirical results analysis.} We further compare the theoretical and empirical memory costs of full fine-tuning in Tab.\,\ref{tab:total_mem_ft_measurement} and LoRA fine-tuning in Tab.\,\ref{tab:total_mem_lora_measurement}. Notably, the empirical results are generally aligned with the theoretical analysis. ZO methods are generally much more efficient than their FO counterparts, and the overhead can be further reduced by loading the half-precision model (using F16).

\begin{figure}[t]
\centering
\includegraphics[width=.9\linewidth]{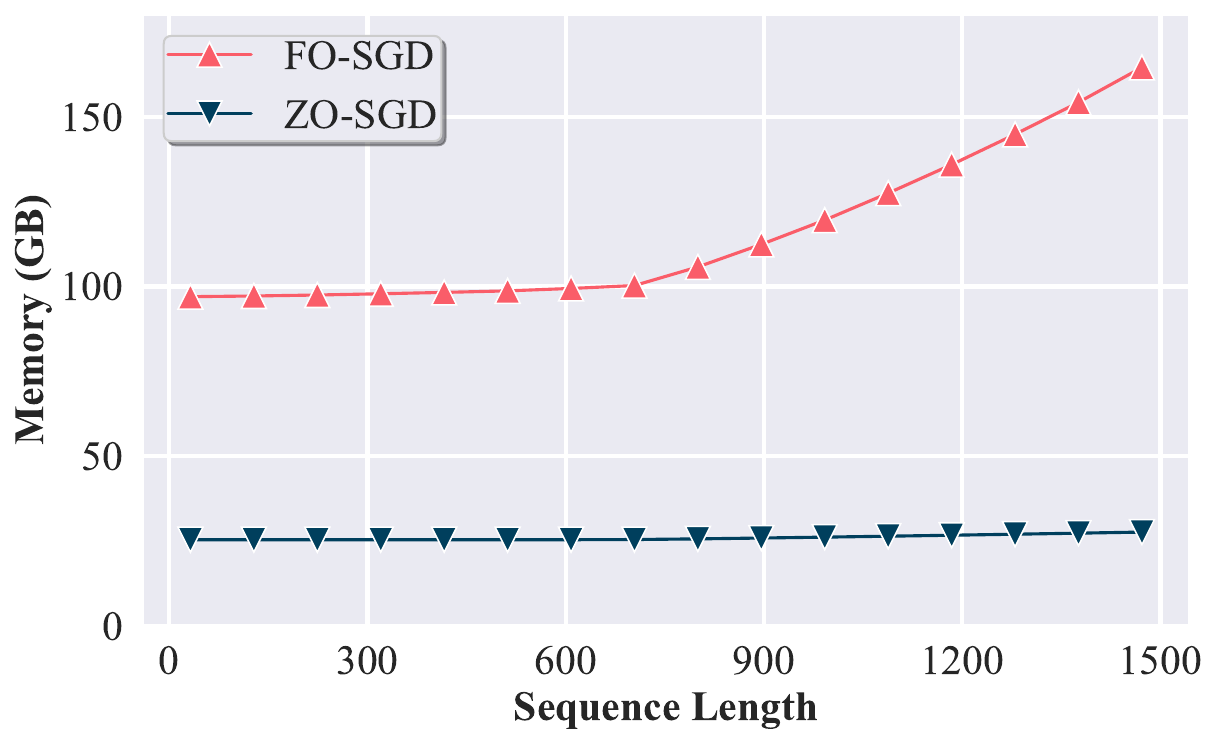}
\vspace{-1em}
\caption{Peak memory comparison of full fine-tuning with FO-SGD and ZO-SGD across various sequence lengths with a fixed effective batch size of $2$. Peak memory consumption was evaluated with the input of synthetic texts generated from random sequences of the specified lengths. 
}
\vspace{-2em}
\label{fig: seqlen_ablation}
\end{figure}

\textbf{A larger sequence length strengthens the memory benefits of ZO methods.} We remark that the empirical results in Tab.\,\ref{tab:total_mem_ft_measurement} and Tab.\,\ref{tab:total_mem_lora_measurement} are dependent on the input sequence length. In general, a larger sequence length will directly lead to a larger activation memory consumption. To investigate its impact, we further explore how the scaling of input sequence length impacts memory usage by increasing activation memory, and the results are shown in Fig. \ref{fig: seqlen_ablation}. Furthermore, we examine the memory cost of LLM fine-tuning vs. the input sequence length. In \textbf{Fig.\,\ref{fig: seqlen_ablation}}, we compare the memory efficiency between ZO-SGD and FO-SGD across various sequence lengths (\textit{i.e.} the token number per sample). As we can see, the memory consumption of ZO-SGD maintains a consistent level, since its peak memory consumption is only determined by the model parameter size (\textit{i.e.}, {the cardinality $|\mathbf x|$}) and independent of the size of the intermediate results (\textit{i.e.}, {$|\mathbf a|$}), see Tab.\,\ref{tab:total_mem_com}. In contrast, as the sequence length increases, the peak memory consumption of FO-SGD first maintains and then begins to increase.
This phenomenon occurs because the peak memory consumption of FO-SGD is determined by the larger one of the per-layer activation storage and the gradient storage (\textit{i.e.}, $\max\left\{|\rva_l|, |\rvx_l|\right\}$). When the sequence length increases, the former begins to grow and eventually exceeds the latter at a certain point (\textit{e.g.}, exceeding 700). Thus, ZO-SGD will exhibit a much better efficiency advantage in settings with long context lengths.

\section{Extended Study to Improve ZO Fine-Tuning}
\label{sec: advanced_zo}

Beyond the benchmarking effort in Sec.\,\ref{sec: benchmark}, we will explore algorithmic advancements to further improve the effectiveness of ZO LLM fine-tuning.
We will leverage the following techniques: 
(1)~\textit{Block-wise ZO fine-tuning}; (2)~\textit{Hybrid ZO and FO fine-tuning}; and  (3) ~\textit{sparsity-induced ZO gradient estimation}. These designs aim to reduce the large variances in gradient estimation when using ZO algorithms.

\noindent\textbf{Block-wise ZO optimization enhances fine-tuning performance.} 
It has been shown in \cite{liu2018zeroth} that using a coordinate-wise deterministic gradient estimator can reduce ZO optimization variance, although this scheme is difficult to scale. In a similar vein, we ask if   \ref{eq: RGE} when estimating a FO gradient \textit{block-wise} can also improve the performance of ZO optimization.
The key idea is to split the LLM into different blocks and then apply the ZO gradient estimator to each block of parameters. For example, {OPT-1.3B} entails $p=26$ parameter blocks, necessitating $p$ forward passes for ZO gradient estimation per fine-tuning step.
Our rationale is that by conducting gradient estimation block-wise, the resulting gradient estimate's variance will be reduced, thereby potentially improving the fine-tuning performance.  

In \textbf{Tab.\,\ref{tab: block-zo-sgd}}, we compare the performance of the ZO fine-tuning baseline MeZO \cite{malladi2024finetuning} (corresponding to ZO-SGD with a query budget of $q =1$ in Sec.\,\ref{sec: benchmark}) with its block-wise \ref{eq: RGE}-based variant, which we term ZO-SGD-Block.
For a fair comparison with ZO-SGD-Block in terms of query complexity, we also present the performance of another variant of ZO-SGD that uses the full model-wise \ref{eq: RGE} but takes the same query number  $q$ as ZO-SGD-Block per iteration.
Notably, ZO-SGD-Block outperforms  ZO-SGD in different query budget settings across different fine-tuning tasks, showing the benefit of \textit{block-wise} ZO tuning.

\begin{table}[t]
\vspace*{-3mm}
\centering
\caption{Performance comparison of OPT-1.3B on the SST2 \& WinoGrande tasks between ZO-SGD and ZO-SGD-Block. The \# of parameter blocks in ZO-SGD-Block is set to $p = 26$. Thus, ZO-SGD w/ the query budget $q=26$ has the same forward pass count as ZO-SGD-Block. MeZO corresponds to ZO-SGD w/ $q = 1$. Best performance for each task is highlighted in \textbf{bold}.}
\label{tab: block-zo-sgd}
\resizebox{.8\linewidth}{!}{
\begin{tabular}{l|c|cc}
\toprule[1pt]
\midrule
Optimizer & Forward Pass \# & SST2 & WinoGrande \\
\midrule
MeZO & $1$   & $90.83$ & $55.5$  \\
ZO-SGD ($q=26$) & $26$  & $91.28$  & $55.7$    \\ 
\midrule
ZO-SGD-Block & $26$  & $\mathbf{93.69}$  & $\mathbf{57.2}$  \\   
\midrule
\bottomrule[1pt]
\end{tabular}
}
\vspace*{-1.7em}
\end{table}

\textbf{Trade-off between performance and memory efficiency via hybrid ZO-FO training.}
The primary source of memory cost in LLM fine-tuning arises from BP, which involves passing gradients from the deep layers to the shallow layers of the model.
To save memory, a potential approach is to confine BP within the deep layers without propagating it to the shallow layers. Moreover, ZO optimization can be employed for the shallow layers without the need for BP. 
The above approach of combining FO optimization for deep layers and ZO optimization for shallow layers results in a \textit{hybrid ZO-FO fine-tuning scheme} for LLMs.


\begin{table}[htb]
\vspace*{-1.5em}
\centering
\caption{The trade-off between memory  cost (in GB) \textit{v.s.} fine-tuning accuracy (\%) using the hybrid ZO-FO training on the {OPT-1.3B} model over the SST2 dataset. The memory or accuracy gap vs. that of the pure ZO-SGD method is noted by $\Delta$.
}
\label{tab: hybrid-fo-zo}
\resizebox{.8\linewidth}{!}{
\begin{tabular}{lrrrr}
\toprule[1pt]
\midrule
\multirow{2}{*}{} & \multicolumn{2}{c}{Memory (GB)} & \multicolumn{2}{c}{Accuracy ($\%$)} \\
\cmidrule(lr){2-3} \cmidrule(lr){4-5}
ZO Layer   \#  & Memory & $\Delta$Memory & Accuracy & $\Delta$Accuracy \\
\midrule
$0$ (FO-SGD) & $24.29$ & $11.07$ & $91.22$ & $1.98$ \\
\midrule
$4$ & $23.33$ & $10.11$ & $91.12$ & $1.88$\\
$8$ & $22.01$ & $8.79$ & $90.79$ & $1.55$ \\
$12$ & $20.43$ & $7.21$ & $89.48$ & $0.24$ \\
$16$ & $18.98$ & $5.76$ & $89.42$ & $0.18$ \\
$20$ & $15.43$ & $2.21$ & $89.27$ & $0.03$ \\
\midrule
$24$ (ZO-SGD) & $13.22$ & $0.00$ & $89.24$ & $0.00$ \\
\midrule
\bottomrule[1pt]
\end{tabular}
}
\vspace*{-.5em}
\end{table}

\textbf{Tab.\,\ref{tab: hybrid-fo-zo}} presents the performance of using the hybrid ZO-FO fine-tuning scheme on ({OPT-1.3B}, SST2). We examine different variants of this hybrid scheme by deciding `\textit{where}' to split between ZO optimization (for shallow layers) and FO optimization (for deep layers).
Suppose that the model consists of $n$ layers, and we designate the first $k \in [1, n]$ layers for ZO optimization, while the remaining $(n-k)$ layers are allocated for FO optimization. The pure ZO optimization approach corresponds to $p = n$.
The results presented in Tab.\,\ref{tab: hybrid-fo-zo} demonstrate that employing ZO optimization on only the first third of the model layers (\textit{i.e.}, $k \leq 8$) can yield performance comparable to that achieved by fully utilizing FO optimization while also reducing memory usage by approximately 10\%. Furthermore, when half of the layers employ ZO optimization (\textit{i.e.}, $k \geq 12$), the performance achieved is similar to that of full ZO fine-tuning.

\noindent\textbf{Gradient pruning benefits performance.} 
We next explore gradient pruning, a technique known for accelerating model training without compromising convergence~\cite{mcdanel2022accelerating}. Our key idea is to induce sparse parameter perturbations for reducing gradient estimation variance in \ref{eq: RGE}. We begin by leveraging magnitude-based pruning \cite{frankle2018lottery,chen2020lottery} to obtain the layer-wise sparsity ratios. We then generate random pruning masks (following these layer-wise sparsity ratios) and apply them to the weight perturbations in \ref{eq: RGE} per ZO fine-tuning step. 
\textbf{Tab.\,\ref{tab: sparse-gradient}} shows the performance of the sparsity-induced ZO gradient estimation in LLM fine-tuning as a function of the overall sparsity ratio. It becomes evident that choosing a moderate sparsity ratio (\textit{e.g.}, $20\%$) can lead to improved performance over the vanilla ZO optimizer, ZO-SGD.

\begin{table}[t]
\vspace*{-3mm}
    \centering
    \caption{Performance of fine-tuning {OPT-1.3B} on COPA \& SST2 datasets using ZO-SGD at different sparsity ratios. A sparsity of $0\%$ represents the baseline, the vanilla ZO-SGD, \textit{i.e.}, MeZO \cite{malladi2024finetuning}. Performance that surpasses the baseline (w/ $0\%$ sparsity) is highlighted in \textbf{bold}.}
    \label{tab: sparse-gradient}
    \resizebox{\linewidth}{!}{
    \begin{tabular}{l|cccccccccc}
    \toprule[1pt]
    \midrule
    \multicolumn{11}{c}{COPA} \\ 
    \midrule
    Sparsity ($\%$) & $0$     & $10$    & $20$    & $30$    & $40$    & $50$    & $60$    & $70$    & $80$    & $90$    \\
    \midrule
    Accuracy ($\%$) & $73.00$    & $\mathbf{75.00}$    & $\mathbf{75.00}$    & $70.00$    & $70.00$    & $70.00$    & $70.00$    & $7.000$    & $70.00$    & $71.00$    \\ 
    \midrule
    \multicolumn{11}{c}{SST2} \\ 
    \hline
    Sparsity ($\%$) & $0$     & $10$    & $20$    & $30$    & $40$    & $50$    & $6 0$    & $70$    & $80$    & $90$    \\
    \midrule
    Accuracy ($\%$) & $90.83$ & $\mathbf{91.51}$ & $\mathbf{92.20}$ & $\mathbf{92.32}$ & $\mathbf{91.74}$ & $\mathbf{92.43}$ & $\mathbf{92.43}$ & $\mathbf{92.20}$ & $\mathbf{91.51}$ & $\mathbf{92.66}$ \\
    \midrule
    \bottomrule[1pt]
    \end{tabular}
    }
    \vspace*{-1.5em}
\end{table}

\section{Conclusion}
\label{sec: conclusion}

This work explores the application of zeroth-order (ZO) optimization in fine-tuning LLMs. ZO optimization approximates gradients using loss differences, eliminating the need for back-propagation and activation storage. While MeZO \cite{malladi2024finetuning} has made strides in adapting ZO optimization for LLMs, understanding the full ZO landscape remains an open question. To address this question, we broaden the scope by considering various ZO optimization methods, task types, and evaluation metrics. We conduct the first benchmark study of different ZO optimization techniques, shedding light on their accuracy and efficiency. 
We also uncover the overlooked ZO optimization principles, such as task alignment and the role of forward gradient. Leveraging these insights, we propose techniques like block-wise descent, hybrid ZO and FO training, and gradient sparsity to enhance ZO optimization-based LLM fine-tuning. 
The proposed enhancements can further improve the fine-tuning accuracy while maintaining the memory efficiency.

\section*{Impact Statement}
This paper aims to advance the optimization foundations of the memory-efficient fine-tuning of large language models (LLMs). Its potential impacts are contingent on how these fine-tuned LLMs are utilized. On the positive side, achieving memory efficiency during LLM fine-tuning could lead to significant reductions in energy consumption, contributing to the development of green AI and achieving improved performance in resource-constrained environments. However, there is a potential negative aspect in terms of misuse, as the fine-tuned models could be employed for generating misinformation, phishing attacks, or releasing copyrighted and private information. However, given the technical focus of this work, there are no specific societal consequences directly stemming from it that need to be highlighted here.

\section*{Acknowledgement}
The work of Y. Zhang and S. Liu is partially supported by the U.S. Department of Energy via Lawrence Livermore National Laboratory. The work of J. D. Lee and Z. Wang is in part supported by the NSF AI Institute for Foundations of Machine Learning (IFML). The work of M. Hong and J. Li is partially supported by NSF grant CCF-1910385 and CNS-2003033. This work is also supported by an Amazon AWS gift award.

\bibliographystyle{icml2024}
\bibliography{ref}

\begin{thebibliography}{87}
\providecommand{\natexlab}[1]{#1}
\providecommand{\url}[1]{\texttt{#1}}
\expandafter\ifx\csname urlstyle\endcsname\relax
  \providecommand{\doi}[1]{doi: #1}\else
  \providecommand{\doi}{doi: \begingroup \urlstyle{rm}\Url}\fi

\bibitem[Amari(1993)]{amari1993backpropagation}
Amari, S.-i.
\newblock Backpropagation and stochastic gradient descent method.
\newblock \emph{Neurocomputing}, 5\penalty0 (4-5):\penalty0 185--196, 1993.

\bibitem[Balasubramanian \& Ghadimi(2018)Balasubramanian and Ghadimi]{balasubramanian2018zeroth}
Balasubramanian, K. and Ghadimi, S.
\newblock Zeroth-order (non)-convex stochastic optimization via conditional gradient and gradient updates.
\newblock \emph{Advances in Neural Information Processing Systems}, 31, 2018.

\bibitem[Baydin et~al.(2022)Baydin, Pearlmutter, Syme, Wood, and Torr]{baydin2022gradients}
Baydin, A.~G., Pearlmutter, B.~A., Syme, D., Wood, F., and Torr, P.
\newblock Gradients without backpropagation.
\newblock \emph{arXiv preprint arXiv:2202.08587}, 2022.

\bibitem[Belouze(2022)]{belouze2022optimization}
Belouze, G.
\newblock Optimization without backpropagation.
\newblock \emph{arXiv preprint arXiv:2209.06302}, 2022.

\bibitem[Boopathy \& Fiete(2022)Boopathy and Fiete]{boopathy2022train}
Boopathy, A. and Fiete, I.
\newblock How to train your wide neural network without backprop: An input-weight alignment perspective.
\newblock In \emph{International Conference on Machine Learning}, pp.\  2178--2205. PMLR, 2022.

\bibitem[Cai et~al.(2021)Cai, Lou, McKenzie, and Yin]{cai2021zeroth}
Cai, H., Lou, Y., McKenzie, D., and Yin, W.
\newblock A zeroth-order block coordinate descent algorithm for huge-scale black-box optimization.
\newblock \emph{arXiv preprint arXiv:2102.10707}, 2021.

\bibitem[Cai et~al.(2022)Cai, Mckenzie, Yin, and Zhang]{cai2022zeroth}
Cai, H., Mckenzie, D., Yin, W., and Zhang, Z.
\newblock Zeroth-order regularized optimization (zoro): Approximately sparse gradients and adaptive sampling.
\newblock \emph{SIAM Journal on Optimization}, 32\penalty0 (2):\penalty0 687--714, 2022.

\bibitem[Chai et~al.(2022)Chai, Wang, Sun, Tian, Wu, and Wang]{chai2022cliptuning}
Chai, Y., Wang, S., Sun, Y., Tian, H., Wu, H., and Wang, H.
\newblock Clip-tuning: Towards derivative-free prompt learning with a mixture of rewards, 2022.

\bibitem[Chen et~al.(2024)Chen, Zhang, Jia, Diffenderfer, Liu, Parasyris, Zhang, Zhang, Kailkhura, and Liu]{chen2023deepzero}
Chen, A., Zhang, Y., Jia, J., Diffenderfer, J., Liu, J., Parasyris, K., Zhang, Y., Zhang, Z., Kailkhura, B., and Liu, S.
\newblock Deepzero: Scaling up zeroth-order optimization for deep model training.
\newblock \emph{ICLR}, 2024.

\bibitem[Chen et~al.(2017)Chen, Zhang, Sharma, Yi, and Hsieh]{chen2017zoo}
Chen, P.-Y., Zhang, H., Sharma, Y., Yi, J., and Hsieh, C.-J.
\newblock Zoo: Zeroth order optimization based black-box attacks to deep neural networks without training substitute models.
\newblock In \emph{Proceedings of the 10th ACM workshop on artificial intelligence and security}, pp.\  15--26, 2017.

\bibitem[Chen et~al.(2022)Chen, Ge, Tong, Wang, Song, Wang, and Luo]{chen2022adaptformer}
Chen, S., Ge, C., Tong, Z., Wang, J., Song, Y., Wang, J., and Luo, P.
\newblock Adaptformer: Adapting vision transformers for scalable visual recognition.
\newblock \emph{Advances in Neural Information Processing Systems}, 35:\penalty0 16664--16678, 2022.

\bibitem[Chen et~al.(2020)Chen, Frankle, Chang, Liu, Zhang, Wang, and Carbin]{chen2020lottery}
Chen, T., Frankle, J., Chang, S., Liu, S., Zhang, Y., Wang, Z., and Carbin, M.
\newblock The lottery ticket hypothesis for pre-trained bert networks.
\newblock \emph{Advances in neural information processing systems}, 33:\penalty0 15834--15846, 2020.

\bibitem[Chen et~al.(2019)Chen, Liu, Xu, Li, Lin, Hong, and Cox]{chen2019zo}
Chen, X., Liu, S., Xu, K., Li, X., Lin, X., Hong, M., and Cox, D.
\newblock Zo-adamm: Zeroth-order adaptive momentum method for black-box optimization.
\newblock \emph{NeurIPS}, 2019.

\bibitem[Cheng et~al.(2021)Cheng, Wu, and Zhu]{cheng2021convergence}
Cheng, S., Wu, G., and Zhu, J.
\newblock On the convergence of prior-guided zeroth-order optimization algorithms.
\newblock \emph{Advances in Neural Information Processing Systems}, 34:\penalty0 14620--14631, 2021.

\bibitem[Deng et~al.(2022)Deng, Wang, Hsieh, Wang, Guo, Shu, Song, Xing, and Hu]{deng2022rlprompt}
Deng, M., Wang, J., Hsieh, C.-P., Wang, Y., Guo, H., Shu, T., Song, M., Xing, E.~P., and Hu, Z.
\newblock Rlprompt: Optimizing discrete text prompts with reinforcement learning, 2022.

\bibitem[Dhurandhar et~al.(2019)Dhurandhar, Pedapati, Balakrishnan, Chen, Shanmugam, and Puri]{dhurandhar2019model}
Dhurandhar, A., Pedapati, T., Balakrishnan, A., Chen, P.-Y., Shanmugam, K., and Puri, R.
\newblock Model agnostic contrastive explanations for structured data.
\newblock \emph{arXiv preprint arXiv:1906.00117}, 2019.

\bibitem[Duchi et~al.(2015)Duchi, Jordan, Wainwright, and Wibisono]{duchi2015optimal}
Duchi, J.~C., Jordan, M.~I., Wainwright, M.~J., and Wibisono, A.
\newblock Optimal rates for zero-order convex optimization: The power of two function evaluations.
\newblock \emph{IEEE Transactions on Information Theory}, 61\penalty0 (5):\penalty0 2788--2806, 2015.

\bibitem[Flaxman et~al.(2004)Flaxman, Kalai, and McMahan]{flaxman2004online}
Flaxman, A.~D., Kalai, A.~T., and McMahan, H.~B.
\newblock Online convex optimization in the bandit setting: gradient descent without a gradient.
\newblock \emph{arXiv preprint cs/0408007}, 2004.

\bibitem[Flaxman et~al.(2005)Flaxman, Kalai, and McMahan]{flaxman2005online}
Flaxman, A.~D., Kalai, A.~T., and McMahan, H.~B.
\newblock Online convex optimization in the bandit setting: {G}radient descent without a gradient.
\newblock In \emph{Proceedings of the sixteenth annual ACM-SIAM symposium on Discrete algorithms}, pp.\  385--394, 2005.

\bibitem[Frankle \& Carbin(2018)Frankle and Carbin]{frankle2018lottery}
Frankle, J. and Carbin, M.
\newblock The lottery ticket hypothesis: Finding sparse, trainable neural networks.
\newblock \emph{arXiv preprint arXiv:1803.03635}, 2018.

\bibitem[Gao et~al.(2020)Gao, Fisch, and Chen]{gao2020making}
Gao, T., Fisch, A., and Chen, D.
\newblock Making pre-trained language models better few-shot learners.
\newblock \emph{arXiv preprint arXiv:2012.15723}, 2020.

\bibitem[Ghadimi \& Lan(2013)Ghadimi and Lan]{ghadimi2013stochastic}
Ghadimi, S. and Lan, G.
\newblock Stochastic first-and zeroth-order methods for nonconvex stochastic programming.
\newblock \emph{SIAM Journal on Optimization}, 23\penalty0 (4):\penalty0 2341--2368, 2013.

\bibitem[Gu et~al.(2021{\natexlab{a}})Gu, Liu, Zhang, Geng, and Huang]{gu2021optimizing}
Gu, B., Liu, G., Zhang, Y., Geng, X., and Huang, H.
\newblock Optimizing large-scale hyperparameters via automated learning algorithm.
\newblock \emph{arXiv preprint arXiv:2102.09026}, 2021{\natexlab{a}}.

\bibitem[Gu et~al.(2021{\natexlab{b}})Gu, Feng, Zhao, Ying, Chen, and Pan]{gu2021efficient}
Gu, J., Feng, C., Zhao, Z., Ying, Z., Chen, R.~T., and Pan, D.~Z.
\newblock Efficient on-chip learning for optical neural networks through power-aware sparse zeroth-order optimization.
\newblock In \emph{Proceedings of the AAAI Conference on Artificial Intelligence}, volume~35, pp.\  7583--7591, 2021{\natexlab{b}}.

\bibitem[Han et~al.(2015)Han, Mao, and Dally]{han2015deep}
Han, S., Mao, H., and Dally, W.~J.
\newblock Deep compression: Compressing deep neural networks with pruning, trained quantization and huffman coding.
\newblock \emph{arXiv preprint arXiv:1510.00149}, 2015.

\bibitem[Hinton(2022)]{hinton2022forward}
Hinton, G.
\newblock The forward-forward algorithm: Some preliminary investigations.
\newblock \emph{arXiv preprint arXiv:2212.13345}, 2022.

\bibitem[Hoffman et~al.(2022)Hoffman, Chenthamarakshan, Wadhawan, Chen, and Das]{hoffman2022optimizing}
Hoffman, S.~C., Chenthamarakshan, V., Wadhawan, K., Chen, P.-Y., and Das, P.
\newblock Optimizing molecules using efficient queries from property evaluations.
\newblock \emph{Nature Machine Intelligence}, 4\penalty0 (1):\penalty0 21--31, 2022.

\bibitem[Hogan \& Kailkhura(2018)Hogan and Kailkhura]{hogan2018universal}
Hogan, T.~A. and Kailkhura, B.
\newblock Universal decision-based black-box perturbations: Breaking security-through-obscurity defenses.
\newblock \emph{arXiv preprint arXiv:1811.03733}, 2018.

\bibitem[Houlsby et~al.(2019)Houlsby, Giurgiu, Jastrzebski, Morrone, De~Laroussilhe, Gesmundo, Attariyan, and Gelly]{houlsby2019parameter}
Houlsby, N., Giurgiu, A., Jastrzebski, S., Morrone, B., De~Laroussilhe, Q., Gesmundo, A., Attariyan, M., and Gelly, S.
\newblock Parameter-efficient transfer learning for nlp.
\newblock In \emph{International Conference on Machine Learning}, pp.\  2790--2799. PMLR, 2019.

\bibitem[Hu et~al.(2021{\natexlab{a}})Hu, Shen, Wallis, Allen-Zhu, Li, Wang, Wang, and Chen]{hu2021lora}
Hu, E.~J., Shen, Y., Wallis, P., Allen-Zhu, Z., Li, Y., Wang, S., Wang, L., and Chen, W.
\newblock Lora: Low-rank adaptation of large language models, 2021{\natexlab{a}}.

\bibitem[Hu et~al.(2021{\natexlab{b}})Hu, Ding, Wang, Liu, Wang, Li, Wu, and Sun]{hu2021knowledgeable}
Hu, S., Ding, N., Wang, H., Liu, Z., Wang, J., Li, J., Wu, W., and Sun, M.
\newblock Knowledgeable prompt-tuning: Incorporating knowledge into prompt verbalizer for text classification.
\newblock \emph{arXiv preprint arXiv:2108.02035}, 2021{\natexlab{b}}.

\bibitem[Huang et~al.(2022)Huang, Gao, Pei, and Huang]{huang2022accelerated}
Huang, F., Gao, S., Pei, J., and Huang, H.
\newblock Accelerated zeroth-order and first-order momentum methods from mini to minimax optimization.
\newblock \emph{The Journal of Machine Learning Research}, 23\penalty0 (1):\penalty0 1616--1685, 2022.

\bibitem[Ilyas et~al.(2018)Ilyas, Engstrom, Athalye, and Lin]{ilyas2018black}
Ilyas, A., Engstrom, L., Athalye, A., and Lin, J.
\newblock Black-box adversarial attacks with limited queries and information.
\newblock In \emph{International conference on machine learning}, pp.\  2137--2146. PMLR, 2018.

\bibitem[Jaderberg et~al.(2017)Jaderberg, Czarnecki, Osindero, Vinyals, Graves, Silver, and Kavukcuoglu]{jaderberg2017decoupled}
Jaderberg, M., Czarnecki, W.~M., Osindero, S., Vinyals, O., Graves, A., Silver, D., and Kavukcuoglu, K.
\newblock Decoupled neural interfaces using synthetic gradients.
\newblock In \emph{International conference on machine learning}, pp.\  1627--1635. PMLR, 2017.

\bibitem[Jiang et~al.(2023)Jiang, Sablayrolles, Mensch, Bamford, Chaplot, Casas, Bressand, Lengyel, Lample, Saulnier, et~al.]{jiang2023mistral}
Jiang, A.~Q., Sablayrolles, A., Mensch, A., Bamford, C., Chaplot, D.~S., Casas, D. d.~l., Bressand, F., Lengyel, G., Lample, G., Saulnier, L., et~al.
\newblock Mistral 7b.
\newblock \emph{arXiv preprint arXiv:2310.06825}, 2023.

\bibitem[Karimi~Mahabadi et~al.(2021)Karimi~Mahabadi, Henderson, and Ruder]{karimi2021compacter}
Karimi~Mahabadi, R., Henderson, J., and Ruder, S.
\newblock Compacter: Efficient low-rank hypercomplex adapter layers.
\newblock \emph{Advances in Neural Information Processing Systems}, 34:\penalty0 1022--1035, 2021.

\bibitem[Khashabi et~al.(2018)Khashabi, Chaturvedi, Roth, Upadhyay, and Roth]{MultiRC2018}
Khashabi, D., Chaturvedi, S., Roth, M., Upadhyay, S., and Roth, D.
\newblock Looking beyond the surface:a challenge set for reading comprehension over multiple sentences.
\newblock In \emph{Proceedings of North American Chapter of the Association for Computational Linguistics (NAACL)}, 2018.

\bibitem[Kim et~al.(2021)Kim, Cai, McKenzie, and Yin]{kim2021curvature}
Kim, B., Cai, H., McKenzie, D., and Yin, W.
\newblock Curvature-aware derivative-free optimization.
\newblock \emph{arXiv preprint arXiv:2109.13391}, 2021.

\bibitem[Kim et~al.(2023)Kim, Kim, Yu, and Chun]{pmlr-v202-kim23l}
Kim, T., Kim, H., Yu, G.-I., and Chun, B.-G.
\newblock {BP}ipe: Memory-balanced pipeline parallelism for training large language models.
\newblock In Krause, A., Brunskill, E., Cho, K., Engelhardt, B., Sabato, S., and Scarlett, J. (eds.), \emph{Proceedings of the 40th International Conference on Machine Learning}, volume 202 of \emph{Proceedings of Machine Learning Research}, pp.\  16639--16653. PMLR, 23--29 Jul 2023.

\bibitem[Kingma \& Ba(2014)Kingma and Ba]{kingma2014adam}
Kingma, D.~P. and Ba, J.
\newblock Adam: A method for stochastic optimization.
\newblock \emph{arXiv preprint arXiv:1412.6980}, 2014.

\bibitem[Lester et~al.(2021)Lester, Al-Rfou, and Constant]{lester2021power}
Lester, B., Al-Rfou, R., and Constant, N.
\newblock The power of scale for parameter-efficient prompt tuning, 2021.

\bibitem[Li \& Liang(2021)Li and Liang]{li2021prefixtuning}
Li, X.~L. and Liang, P.
\newblock Prefix-tuning: Optimizing continuous prompts for generation, 2021.

\bibitem[Lin et~al.(2020)Lin, Madotto, and Fung]{lin2020exploring}
Lin, Z., Madotto, A., and Fung, P.
\newblock Exploring versatile generative language model via parameter-efficient transfer learning, 2020.

\bibitem[Liu et~al.(2018)Liu, Kailkhura, Chen, Ting, Chang, and Amini]{liu2018zeroth}
Liu, S., Kailkhura, B., Chen, P.-Y., Ting, P., Chang, S., and Amini, L.
\newblock Zeroth-order stochastic variance reduction for nonconvex optimization.
\newblock volume~31, 2018.

\bibitem[Liu et~al.(2019{\natexlab{a}})Liu, Chen, Chen, and Hong]{liu2018signsgd}
Liu, S., Chen, P.-Y., Chen, X., and Hong, M.
\newblock sign{SGD} via zeroth-order oracle.
\newblock In \emph{International Conference on Learning Representations}, 2019{\natexlab{a}}.

\bibitem[Liu et~al.(2020)Liu, Chen, Kailkhura, Zhang, Hero~III, and Varshney]{liu2020primer}
Liu, S., Chen, P.-Y., Kailkhura, B., Zhang, G., Hero~III, A.~O., and Varshney, P.~K.
\newblock A primer on zeroth-order optimization in signal processing and machine learning: Principals, recent advances, and applications.
\newblock volume~37, pp.\  43--54. IEEE, 2020.

\bibitem[Liu et~al.(2021)Liu, Ji, Fu, Tam, Du, Yang, and Tang]{liu2021p}
Liu, X., Ji, K., Fu, Y., Tam, W.~L., Du, Z., Yang, Z., and Tang, J.
\newblock P-tuning v2: Prompt tuning can be comparable to fine-tuning universally across scales and tasks.
\newblock \emph{arXiv preprint arXiv:2110.07602}, 2021.

\bibitem[Liu et~al.(2022)Liu, Ji, Fu, Tam, Du, Yang, and Tang]{liu2022p}
Liu, X., Ji, K., Fu, Y., Tam, W., Du, Z., Yang, Z., and Tang, J.
\newblock P-tuning: Prompt tuning can be comparable to fine-tuning across scales and tasks.
\newblock In \emph{Proceedings of the 60th Annual Meeting of the Association for Computational Linguistics (Volume 2: Short Papers)}, pp.\  61--68, 2022.

\bibitem[Liu et~al.(2019{\natexlab{b}})Liu, Ott, Goyal, Du, Joshi, Chen, Levy, Lewis, Zettlemoyer, and Stoyanov]{liu2019roberta}
Liu, Y., Ott, M., Goyal, N., Du, J., Joshi, M., Chen, D., Levy, O., Lewis, M., Zettlemoyer, L., and Stoyanov, V.
\newblock Roberta: A robustly optimized bert pretraining approach.
\newblock \emph{arXiv preprint arXiv:1907.11692}, 2019{\natexlab{b}}.

\bibitem[Luo et~al.(2023)Luo, Huang, Zhou, Sun, Jiang, Wang, and Ji]{luo2023towards}
Luo, G., Huang, M., Zhou, Y., Sun, X., Jiang, G., Wang, Z., and Ji, R.
\newblock Towards efficient visual adaption via structural re-parameterization.
\newblock \emph{arXiv preprint arXiv:2302.08106}, 2023.

\bibitem[Malladi et~al.(2023)Malladi, Gao, Nichani, Damian, Lee, Chen, and Arora]{malladi2024finetuning}
Malladi, S., Gao, T., Nichani, E., Damian, A., Lee, J.~D., Chen, D., and Arora, S.
\newblock Fine-tuning language models with just forward passes.
\newblock \emph{arXiv preprint arXiv:2305.17333}, 2023.

\bibitem[McDanel et~al.(2022)McDanel, Dinh, and Magallanes]{mcdanel2022accelerating}
McDanel, B., Dinh, H., and Magallanes, J.
\newblock Accelerating dnn training with structured data gradient pruning.
\newblock In \emph{2022 26th International Conference on Pattern Recognition (ICPR)}, pp.\  2293--2299. IEEE, 2022.

\bibitem[Meier et~al.(2019)Meier, Mujika, Gauy, and Steger]{meier2019improving}
Meier, F., Mujika, A., Gauy, M.~M., and Steger, A.
\newblock Improving gradient estimation in evolutionary strategies with past descent directions.
\newblock \emph{arXiv preprint arXiv:1910.05268}, 2019.

\bibitem[Nesterov \& Spokoiny(2017)Nesterov and Spokoiny]{nesterov2017random}
Nesterov, Y. and Spokoiny, V.
\newblock Random gradient-free minimization of convex functions.
\newblock \emph{Foundations of Computational Mathematics}, 17:\penalty0 527--566, 2017.

\bibitem[N{\o}kland \& Eidnes(2019)N{\o}kland and Eidnes]{nokland2019training}
N{\o}kland, A. and Eidnes, L.~H.
\newblock Training neural networks with local error signals.
\newblock In \emph{International conference on machine learning}, pp.\  4839--4850. PMLR, 2019.

\bibitem[Ohta et~al.(2020)Ohta, Berger, Sokolov, and Riezler]{ohta2020sparse}
Ohta, M., Berger, N., Sokolov, A., and Riezler, S.
\newblock Sparse perturbations for improved convergence in stochastic zeroth-order optimization.
\newblock In \emph{Machine Learning, Optimization, and Data Science: 6th International Conference, LOD 2020, Siena, Italy, July 19--23, 2020, Revised Selected Papers, Part II 6}, pp.\  39--64. Springer, 2020.

\bibitem[Pfeiffer et~al.(2020)Pfeiffer, R{\"u}ckl{\'e}, Poth, Kamath, Vuli{\'c}, Ruder, Cho, and Gurevych]{pfeiffer2020adapterhub}
Pfeiffer, J., R{\"u}ckl{\'e}, A., Poth, C., Kamath, A., Vuli{\'c}, I., Ruder, S., Cho, K., and Gurevych, I.
\newblock Adapterhub: A framework for adapting transformers.
\newblock \emph{arXiv preprint arXiv:2007.07779}, 2020.

\bibitem[Prasad et~al.(2023)Prasad, Hase, Zhou, and Bansal]{prasad2023grips}
Prasad, A., Hase, P., Zhou, X., and Bansal, M.
\newblock Grips: Gradient-free, edit-based instruction search for prompting large language models, 2023.

\bibitem[Raffel et~al.(2023)Raffel, Shazeer, Roberts, Lee, Narang, Matena, Zhou, Li, and Liu]{raffel2023exploring}
Raffel, C., Shazeer, N., Roberts, A., Lee, K., Narang, S., Matena, M., Zhou, Y., Li, W., and Liu, P.~J.
\newblock Exploring the limits of transfer learning with a unified text-to-text transformer, 2023.

\bibitem[Ren et~al.(2021)Ren, Rajbhandari, Aminabadi, Ruwase, Yang, Zhang, Li, and He]{ren2021zerooffload}
Ren, J., Rajbhandari, S., Aminabadi, R.~Y., Ruwase, O., Yang, S., Zhang, M., Li, D., and He, Y.
\newblock Zero-offload: Democratizing billion-scale model training, 2021.

\bibitem[Ren et~al.(2022)Ren, Kornblith, Liao, and Hinton]{ren2022scaling}
Ren, M., Kornblith, S., Liao, R., and Hinton, G.
\newblock Scaling forward gradient with local losses.
\newblock \emph{arXiv preprint arXiv:2210.03310}, 2022.

\bibitem[Roemmele et~al.(2011)Roemmele, Bejan, and Gordon]{roemmele2011choice}
Roemmele, M., Bejan, C.~A., and Gordon, A.~S.
\newblock Choice of plausible alternatives: An evaluation of commonsense causal reasoning.
\newblock In \emph{2011 AAAI Spring Symposium Series}, 2011.

\bibitem[Sakaguchi et~al.(2021)Sakaguchi, Bras, Bhagavatula, and Choi]{sakaguchi2021winogrande}
Sakaguchi, K., Bras, R.~L., Bhagavatula, C., and Choi, Y.
\newblock Winogrande: An adversarial winograd schema challenge at scale.
\newblock \emph{Communications of the ACM}, 64\penalty0 (9):\penalty0 99--106, 2021.

\bibitem[Sanh et~al.(2022)Sanh, Webson, Raffel, Bach, Sutawika, Alyafeai, Chaffin, Stiegler, Scao, Raja, Dey, Bari, Xu, Thakker, Sharma, Szczechla, Kim, Chhablani, Nayak, Datta, Chang, Jiang, Wang, Manica, Shen, Yong, Pandey, Bawden, Wang, Neeraj, Rozen, Sharma, Santilli, Fevry, Fries, Teehan, Bers, Biderman, Gao, Wolf, and Rush]{sanh2022multitask}
Sanh, V., Webson, A., Raffel, C., Bach, S.~H., Sutawika, L., Alyafeai, Z., Chaffin, A., Stiegler, A., Scao, T.~L., Raja, A., Dey, M., Bari, M.~S., Xu, C., Thakker, U., Sharma, S.~S., Szczechla, E., Kim, T., Chhablani, G., Nayak, N., Datta, D., Chang, J., Jiang, M. T.-J., Wang, H., Manica, M., Shen, S., Yong, Z.~X., Pandey, H., Bawden, R., Wang, T., Neeraj, T., Rozen, J., Sharma, A., Santilli, A., Fevry, T., Fries, J.~A., Teehan, R., Bers, T., Biderman, S., Gao, L., Wolf, T., and Rush, A.~M.
\newblock Multitask prompted training enables zero-shot task generalization, 2022.

\bibitem[Shu et~al.(2022)Shu, Dai, Sng, Verma, Jaillet, and Low]{shu2022zeroth}
Shu, Y., Dai, Z., Sng, W., Verma, A., Jaillet, P., and Low, B. K.~H.
\newblock Zeroth-order optimization with trajectory-informed derivative estimation.
\newblock In \emph{The Eleventh International Conference on Learning Representations}, 2022.

\bibitem[Silver et~al.(2021)Silver, Goyal, Danihelka, Hessel, and van Hasselt]{silver2021learning}
Silver, D., Goyal, A., Danihelka, I., Hessel, M., and van Hasselt, H.
\newblock Learning by directional gradient descent.
\newblock In \emph{International Conference on Learning Representations}, 2021.

\bibitem[Singhal et~al.(2023)Singhal, Cheung, Chandra, Ragan-Kelley, Tenenbaum, Poggio, and Yu]{singhal2023guess}
Singhal, U., Cheung, B., Chandra, K., Ragan-Kelley, J., Tenenbaum, J.~B., Poggio, T.~A., and Yu, S.~X.
\newblock How to guess a gradient.
\newblock \emph{arXiv preprint arXiv:2312.04709}, 2023.

\bibitem[Socher et~al.(2013)Socher, Perelygin, Wu, Chuang, Manning, Ng, and Potts]{socher2013recursive}
Socher, R., Perelygin, A., Wu, J., Chuang, J., Manning, C.~D., Ng, A.~Y., and Potts, C.
\newblock Recursive deep models for semantic compositionality over a sentiment treebank.
\newblock In \emph{Proceedings of the 2013 conference on empirical methods in natural language processing}, pp.\  1631--1642, 2013.

\bibitem[Sun et~al.(2022{\natexlab{a}})Sun, He, Qian, Zhou, Huang, and Qiu]{sun2022bbtv2}
Sun, T., He, Z., Qian, H., Zhou, Y., Huang, X., and Qiu, X.
\newblock Bbtv2: Towards a gradient-free future with large language models, 2022{\natexlab{a}}.

\bibitem[Sun et~al.(2022{\natexlab{b}})Sun, Shao, Qian, Huang, and Qiu]{sun2022black}
Sun, T., Shao, Y., Qian, H., Huang, X., and Qiu, X.
\newblock Black-box tuning for language-model-as-a-service.
\newblock In \emph{International Conference on Machine Learning}, pp.\  20841--20855. PMLR, 2022{\natexlab{b}}.

\bibitem[Tan et~al.(2021)Tan, Zhang, Wang, and Liu]{tan2021msp}
Tan, Z., Zhang, X., Wang, S., and Liu, Y.
\newblock Msp: Multi-stage prompting for making pre-trained language models better translators.
\newblock \emph{arXiv preprint arXiv:2110.06609}, 2021.

\bibitem[Touvron et~al.(2023)Touvron, Martin, Stone, Albert, Almahairi, Babaei, Bashlykov, Batra, Bhargava, Bhosale, et~al.]{touvron2023llama}
Touvron, H., Martin, L., Stone, K., Albert, P., Almahairi, A., Babaei, Y., Bashlykov, N., Batra, S., Bhargava, P., Bhosale, S., et~al.
\newblock Llama 2: Open foundation and fine-tuned chat models.
\newblock \emph{arXiv preprint arXiv:2307.09288}, 2023.

\bibitem[Tsai et~al.(2020)Tsai, Chen, and Ho]{tsai2020transfer}
Tsai, Y.-Y., Chen, P.-Y., and Ho, T.-Y.
\newblock Transfer learning without knowing: Reprogramming black-box machine learning models with scarce data and limited resources.
\newblock In \emph{International Conference on Machine Learning}, pp.\  9614--9624. PMLR, 2020.

\bibitem[Tsaknakis et~al.(2022)Tsaknakis, Kailkhura, Liu, Loveland, Diffenderfer, Hiszpanski, and Hong]{tsaknakis2022zeroth}
Tsaknakis, I., Kailkhura, B., Liu, S., Loveland, D., Diffenderfer, J., Hiszpanski, A.~M., and Hong, M.
\newblock Zeroth-order sciml: Non-intrusive integration of scientific software with deep learning.
\newblock \emph{arXiv preprint arXiv:2206.02785}, 2022.

\bibitem[Tu et~al.(2019)Tu, Ting, Chen, Liu, Zhang, Yi, Hsieh, and Cheng]{tu2019autozoom}
Tu, C.-C., Ting, P., Chen, P.-Y., Liu, S., Zhang, H., Yi, J., Hsieh, C.-J., and Cheng, S.-M.
\newblock Autozoom: Autoencoder-based zeroth order optimization method for attacking black-box neural networks.
\newblock In \emph{Proceedings of the AAAI Conference on Artificial Intelligence}, pp.\  742--749, 2019.

\bibitem[Vemula et~al.(2019)Vemula, Sun, and Bagnell]{vemula2019contrasting}
Vemula, A., Sun, W., and Bagnell, J.
\newblock Contrasting exploration in parameter and action space: A zeroth-order optimization perspective.
\newblock In \emph{The 22nd International Conference on Artificial Intelligence and Statistics}, pp.\  2926--2935. PMLR, 2019.

\bibitem[Verma et~al.(2023)Verma, Bangar, Subramanyam, Lal, Shah, and Satoh]{verma2023certified}
Verma, A., Bangar, S., Subramanyam, A., Lal, N., Shah, R.~R., and Satoh, S.
\newblock Certified zeroth-order black-box defense with robust unet denoiser.
\newblock \emph{arXiv preprint arXiv:2304.06430}, 2023.

\bibitem[Vicol et~al.(2023)Vicol, Kolter, and Swersky]{vicol2023low}
Vicol, P., Kolter, Z., and Swersky, K.
\newblock Low-variance gradient estimation in unrolled computation graphs with es-single.
\newblock \emph{arXiv preprint arXiv:2304.11153}, 2023.

\bibitem[Wang et~al.(2019)Wang, Singh, Michael, Hill, Levy, and Bowman]{wang2019glue}
Wang, A., Singh, A., Michael, J., Hill, F., Levy, O., and Bowman, S.~R.
\newblock {GLUE}: A multi-task benchmark and analysis platform for natural language understanding.
\newblock 2019.
\newblock In the Proceedings of ICLR.

\bibitem[Wang et~al.(2022)Wang, Guo, Su, Yang, and Yan]{wang2022zarts}
Wang, X., Guo, W., Su, J., Yang, X., and Yan, J.
\newblock Zarts: On zero-order optimization for neural architecture search.
\newblock \emph{Advances in Neural Information Processing Systems}, 35:\penalty0 12868--12880, 2022.

\bibitem[Wang et~al.(2017)Wang, Du, Balakrishnan, and Singh]{wang2017stochastic}
Wang, Y., Du, S., Balakrishnan, S., and Singh, A.
\newblock Stochastic zeroth-order optimization in high dimensions.
\newblock \emph{arXiv preprint arXiv:1710.10551}, 2017.

\bibitem[Ye et~al.(2018)Ye, Huang, Fang, Li, and Zhang]{ye2018hessian}
Ye, H., Huang, Z., Fang, C., Li, C.~J., and Zhang, T.
\newblock Hessian-aware zeroth-order optimization for black-box adversarial attack.
\newblock \emph{arXiv preprint arXiv:1812.11377}, 2018.

\bibitem[Zhang et~al.(2022{\natexlab{a}})Zhang, Roller, Goyal, Artetxe, Chen, Chen, Dewan, Diab, Li, Lin, et~al.]{zhang2022opt}
Zhang, S., Roller, S., Goyal, N., Artetxe, M., Chen, M., Chen, S., Dewan, C., Diab, M., Li, X., Lin, X.~V., et~al.
\newblock Opt: Open pre-trained transformer language models.
\newblock \emph{arXiv preprint arXiv:2205.01068}, 2022{\natexlab{a}}.

\bibitem[Zhang et~al.(2022{\natexlab{b}})Zhang, Yao, Jia, Yi, Hong, Chang, and Liu]{zhang2022robustify}
Zhang, Y., Yao, Y., Jia, J., Yi, J., Hong, M., Chang, S., and Liu, S.
\newblock How to robustify black-box ml models? a zeroth-order optimization perspective.
\newblock \emph{ICLR}, 2022{\natexlab{b}}.

\bibitem[Zhao et~al.(2019)Zhao, Liu, Chen, Hoang, Xu, Kailkhura, and Lin]{zhao2019design}
Zhao, P., Liu, S., Chen, P.-Y., Hoang, N., Xu, K., Kailkhura, B., and Lin, X.
\newblock On the design of black-box adversarial examples by leveraging gradient-free optimization and operator splitting method.
\newblock In \emph{Proceedings of the IEEE/CVF International Conference on Computer Vision}, pp.\  121--130, 2019.

\bibitem[Zheng et~al.(2023)Zheng, Chiang, Sheng, Zhuang, Wu, Zhuang, Lin, Li, Li, Xing, et~al.]{zheng2023judging}
Zheng, L., Chiang, W.-L., Sheng, Y., Zhuang, S., Wu, Z., Zhuang, Y., Lin, Z., Li, Z., Li, D., Xing, E., et~al.
\newblock Judging llm-as-a-judge with mt-bench and chatbot arena.
\newblock \emph{arXiv preprint arXiv:2306.05685}, 2023.

\bibitem[Zhu et~al.(2023)Zhu, Voigt, Ko, and Rahimian]{zhu2023ondevice}
Zhu, S., Voigt, T., Ko, J., and Rahimian, F.
\newblock On-device training: A first overview on existing systems, 2023.

\end{thebibliography}

\clearpage
\section*{\Large{Appendix}}
\setcounter{section}{0}
\setcounter{figure}{0}
\setcounter{table}{0}
\makeatletter 
\renewcommand{\thesection}{\Alph{section}}
\renewcommand{\theHsection}{\Alph{section}}
\renewcommand{\thefigure}{A\arabic{figure}}
\renewcommand{\theHfigure}{A\arabic{figure}}
\renewcommand{\thetable}{A\arabic{table}}
\renewcommand{\theHtable}{A\arabic{table}}
\makeatother

\renewcommand{\thetable}{A\arabic{table}}
\setcounter{mylemma}{0}
\renewcommand{\themylemma}{A\arabic{mylemma}}
\setcounter{algorithm}{0}
\renewcommand{\thealgorithm}{A\arabic{algorithm}}
\setcounter{equation}{0}
\renewcommand{\theequation}{A\arabic{equation}}

\section{Zeroth-Order Optimization Algorithms}
\label{sec: ZO}

Zeroth-order optimization addresses the minimization or maximization of an objective function $f: \mathbb{R}^n \rightarrow \mathbb{R}$ without the use of derivatives:
\[\min_{\mathbf x \in \mathbb{R}^n} f(\mathbf x)\]
These methods are pivotal when the function is non-differentiable, gradient computation is expensive, or the function evaluations are noisy. Random gradient estimation~(RGE) provides a surrogate for gradients in zeroth-order optimization by sampling function evaluations. The gradient $\nabla f(\mathbf x)$ at a point $\mathbf x$ can be approximated as:
\[\hat{\nabla}f(\mathbf x) := \frac{f(\mathbf x + \mu \mathbf u) - f(\mathbf x - \mu \mathbf u)}{2\mu} \cdot \mathbf u\]
where $\mathbf u$ is a standard Gaussian vector, and $\mu$ is a small scalar. This estimation facilitates the use of gradient-based methods solely based on function evaluations. Utilizing this gradient estimator, we summarize the existing zeroth-other algorithms as follows

$\rhd$ \textit{ZO-SGD.}~\cite{ghadimi2013stochastic}
The method directly update the parameters by the estimated gradient:
\[\mathbf x_{t+1} =\mathbf x_{t} - \eta_t  \hat{\nabla}f(\mathbf x_{t}).\]

$\rhd$ \textit{ZO-Sign-SGD.}~\cite{liu2018signsgd}
The intuition of ZO-Sign-SGD is to make the gradient estimation more robust to noise since the sign operation could mitigate the negative effect of (coordinate-wise) gradient noise of large variance. ZO-Sign-SGD uses two-side RGE and update the parameters as follows,
\[\mathbf x_{t+1} =\mathbf x_{t} - \eta_t  \sign(\frac{f(\mathbf x + \mu \mathbf u) - f(\mathbf x - \mu \mathbf u)}{2\mu})\mathbf u\]
where $u$ is a standard Gaussian vector.

$\rhd$ \textit{ZO-SGD with Momentum.}~\cite{huang2022accelerated}
The stochastic gradient estimated from a batch may suffer from a large variance.
Momentum uses a moving average to estimate the global gradient and update the parameters by
\[\mathbf x_{t+1} =\mathbf x_{t} - \eta_t  \mathbf{m}_t, \]
with the momentum defined as
\[\mathbf{m}_{t} = \beta_t\mathbf{m}_{t-1}+\hat{\nabla}f(\mathbf x_{{t}}).\]

$\rhd$ \textit{ZO-SGD with Conservative Gradient Update.}~\cite{kim2021curvature}
This method is adapted from~\cite{kim2021curvature} to update the parameter in a conservative way: we pick up the point that corresponds to the smallest loss value. The update writes:
\[\mathbf x_{t+1} =\argmin_{\mathbf{y}\in\{\mathbf x_{t}, \mathbf x_{t} - \eta_t  \hat{\nabla}f(\mathbf x_{t}), \mathbf{x}_{t} + \eta_t  \hat{\nabla}f(\mathbf x_{t})\}} f(\mathbf{y})\] 
Due to the large variance introduced by the zeroth-order RGE estimator, we believe that the conservative update could correct the wrong step made by ZO-SGD, yielding better convergence results.

$\rhd$ \textit{ZO-Adam.}~\cite{chen2019zo}
Similar to the ZO-SGD with momentum, ZO-Adam uses momentum to estimate the gradients. In addition, ZO-Adam adaptively penalizes the learning rate by a diagonal matrix $\mathbf{V}_t$ to reduce the noise.
In summary, the parameters will be updated by
\[\mathbf x_{t+1} =\mathbf x_{t} - \eta_t \mathbf{V}_t^{-1/2} \mathbf{m}_t,\]
where the momentum $\mathbf{m}_t$, second raw momentum estimate $\mathbf{v}$ and normalization matrix $\mathbf{V}$ are defined as
\begin{align*}
    \mathbf{m}_{t} &= \beta_{1,t} \mathbf{m}_{t-1}+ (1-\beta_{1,t}) \hat{\nabla}f(\mathbf{x}_t), \\
    \mathbf{v}_{t} &= \beta_2 \mathbf{v}_{t-1} + (1-\beta_2) [\hat{\nabla}f(\mathbf{x}_t)]^2, \\
    \mathbf{V}_{t} &= \text{Diag} (\max (\mathbf{v}_{t}, \mathbf{v}_{t-1})).
\end{align*}

$\rhd$ \textit{Forward Gradient.}\cite{baydin2022gradients}
Forward Gradient was proposed to get unbiased stochastic estimation of directional derivatives.
The update is formulated as
\[\mathbf x_{t+1} =\mathbf x_{t} - \eta_t  (\nabla f(\mathbf x_{t})^\top \mathbf u ) \mathbf u\],
where $\mathbf u$ is the standard Gaussian random vector.
Though the Forward Gradient uses the first-order gradient $\nabla f(\mathbf x_{t})$, it uses the Jacobian-Vector Product (JVP) during the forward pass of training to reduce the computation and memory consumption.
Therefore, the memory complexity is still comparably low compared to back-propagation-based approaches.

\section{Preliminaries of Parameter-Efficient Fine-Tuning (PEFT)} \label{sec:pre_peft}

In our benchmark, we consider three PEFT methods, including \{LoRA, prefix~tuning, prompt~tuning\}. \\
\emph{$1$) Low-Rank Adaptation (LoRA).}
LoRA modifies a pre-trained model by introducing trainable low-rank matrices, enabling fine-tuning with a limited parameter budget. Given a weight matrix $\mathbf W \in \mathbb{R}^{m \times n}$ in a transformer model, LoRA decomposes it as:
\begin{equation}
    \mathbf W' = \mathbf W + \mathbf B\mathbf A
\end{equation}
where $W$ is the original weight matrix, $\mathbf B \in \mathbb{R}^{m \times r}$ and $\mathbf A \in \mathbb{R}^{r \times n}$ are the low-rank matrices, and $r \ll \min(m, n)$ represents the rank. During fine-tuning, only $\mathbf B$ and $\mathbf A$ are updated, keeping $\mathbf W$ frozen. \\
\emph{$2$) Prompt Tuning.}
Prompt tuning introduces a series of trainable tokens, or prompts, to guide the model's predictions. Let $\mathbf x$ be the input sequence and $\mathbf P \in \mathbb{R}^{l \times d}$ the matrix representing the prompt embeddings, where $l$ is the length of the prompt and $d$ is the embedding dimension. The model input is then:
\begin{equation}
    \hat{\mathbf x} = [\mathbf P; \mathbf E(\mathbf x)]
\end{equation}
where $\mathbf E(\mathbf x)$ is the embedding of the original input $\mathbf x$, and $\hat{\mathbf x}$ represents the concatenated input of the prompt and the original input. During fine-tuning, only the prompt embeddings $\mathbf P$ are learned, with the rest of the model parameters kept frozen. \\
\emph{$3$) Prefix Tuning.}
Prefix tuning extends the idea of prompt tuning to the attention mechanism of transformer models. Given an input sequence $\mathbf x$, the model processes it with additional context vectors $\mathbf C_k$ and $\mathbf C_v$ serving as keys and values in the attention mechanism:
\begin{equation}
    \text{Attention}(\mathbf Q, \mathbf K, \mathbf V) = \text{softmax}\left(\frac{\mathbf Q(\mathbf K + \mathbf C_k)^\text{T}}{\sqrt{d_k}}\right)(\mathbf V + \mathbf C_v)
\end{equation}
where $\mathbf Q$, $\mathbf K$ and $\mathbf V$ represent the query, key, and value matrices in the attention mechanism, $\mathbf C_k \in \mathbb{R}^{l \times d_k}$, $\mathbf C_v \in \mathbb{R}^{l \times d_v}$, and $l$ is the length of the prefix. During training, only $\mathbf C_k$ and $\mathbf C_v$ are updated, and the original model parameters are frozen.

\section{How to Implement Memory-Efficient ZO/FO Optimizers?}

\begin{algorithm}[ht]
\caption{A General Pipeline for A FO/ZO Optimizer}
\label{alg:opt}
\begin{algorithmic}
\STATE \textbf{Input:} Model with forward function $f$, parameters $\rvx$, previous optimizer state $\rvs_{\text{opt}}$
\STATE \textbf{Memory Overview:} $\mathbf{s}$: States of optimizer/model, $\boldsymbol\tau$: Temporary variables (\textit{e.g.}, for copy/paste)

\STATE \textbf{Step 0: Model Loading}
\STATE $\bullet$ \textit{Function:} Initialize the model with parameter $\rvx$;
\STATE $\bullet$ \textit{Memory Consumption:} $\rvx$;
\STATE $\bullet$ \textit{Memory Release:} N/A.

\STATE \textbf{Step 1: Forward Pass}
\STATE $\bullet$ \textit{Function:} Compute loss $\ell(x)$, save forward pass states $\rvs_{\text{fwd}}$, and use temporary variables $\btau_{\text{fwd}}$.
\STATE $\bullet$ \textit{Memory Consumption:} $\ell(x), \rvs_{\text{fwd}}, \btau_{\text{fwd}}$;
\STATE $\bullet$ \textit{Memory Release:} Release $\btau_{\text{fwd}}$ after computation.

\STATE \textbf{Step 2: Backward Pass}
\STATE $\bullet$ \textit{Function:} Calculate gradients \textit{w.r.t.} $\rvx$, generate backward states $\rvs_{\text{bwd}}$, employ temporary variables $\btau_{\text{bwd}}$;
\STATE $\bullet$ \textit{Memory Consumption:} $\rvs_{\text{bwd}}, \btau_{\text{bwd}}$;
\STATE $\bullet$ \textit{Memory Release:} Release $\rvs_{\text{fwd}}, \btau_{\text{bwd}}$ after gradients are computed.

\STATE \textbf{Step 3: Optimization Step}
\STATE $\bullet$ \textit{Function:} Update $\rvx$ and $\rvs_{\text{opt}}$ using gradients, utilize temporary variables $\btau_{\text{opt}}$.
\STATE $\bullet$ \textit{Memory Consumption:} $\rvx, \rvs_{\text{opt}}, \btau_{\text{opt}}$.
\STATE $\bullet$ \textit{Memory Release:} Release $\rvs_{\text{bwd}}, \btau_{\text{opt}}$ post-update.

\STATE \textbf{Output:} Loss $\ell$, updated parameters $\rvx$, and optimizer state $\rvs_{\text{opt}}$.
\end{algorithmic}
\end{algorithm}

The memory efficiency of an optimizer heavily depends on its implementation details. In this section, we will discuss these implementation details and provide an holistic memory profile of all the optimizers discussed above. We remark that the discussions in this section are based on the PyTorch framework.
In general, the memory efficiency of an optimizer is defined by the peak memories consumed during training in a specific setting, which primarily depends on the maximum number of variables stored at a certain time point.
To dissect the memory consumption of different optimizers, we summarized a general model parameter updating pipeline in \cref{alg:opt}, which involves four main steps.
\underline{First}, the program needs to load the model with the full parameter $\rvx$ with either full or half-precision. This is an inevitble consumption for all optimizers.
\underline{Second}, a forward pass will yield a loss value and store any involved intermediate state of the model $\rvs_{\text{fwd}}$. This process may also requires storing any other temporary variables $\btau_{\text{fwd}}$ (\textit{e.g.} an extra copy in float16 of model weight in FO-SGD FP16), which will be immediately released after the forward is complete.
\underline{Third}, the loss is back-propagated in a backward mode utilizing the stored state $\rvs_{\text{fwd}}$. Similarly, this process involves storing the backward states $\rvs_{\text{bwd}}$ and temporarily storing $\btau_{\text{bwd}}$.
After the gradients and state $\rvs_{\text{bwd}}$ are computed, the memory cost from $\rvs_{\text{fwd}}$ will be immediately released.
\underline{Last}, with the computed gradient and states, the parameters and optimizer state will be updated, and the memory of $\rvs_{\text{bwd}}$ will be released.
Typically, $\rvs_{\text{bwd}}$ will be the stored gradients.
Since the forward states and gradients are computed layer-wise, the memory consumption will be sequentially transmitted from the state to the gradient.
According to the life-cycle, the memory consumption will be summarized as two types: \emph{constant memory} that exists through the training life-cycle, and \emph{dynamic allocation} that only exists in one iteration mainly for gradient computing.
Though dynamic allocation only temporarily exist, it can increase the peak memory.
As a result, we summarize the peak memory consumption as
{
\small
\begin{align*}
    \underbrace{ |\rvx| + |\rvs_{\text{opt}}| }_{\text{constant}} + \underbrace{ \max \left\{|\rvs_{\text{fwd}}|  + |\btau_{\text{fwd}}|, |\rvs_{\text{bwd}}| + |\btau_{\text{bwd}}|, |\rvs_{\text{bwd}}| + |\btau_{\text{opt}}| \right\} }_{\text{dynamic allocation}}.
\end{align*}}%

If the forward and backward are computed per layer and the temporary states can be voided, then the term $\displaystyle \max\{|\rvs_{\text{fwd}}| + |\btau_{\text{fwd}}|, |\rvs_{\text{bwd}}| + |\btau_{\text{bwd}}|\}$ will be replaced by $\displaystyle \sum_l \max\{|\rvs_{\text{fwd},l}|, |\rvs_{\text{bwd},l}|\}$, where $l$ represents any layer number, leading to a memory consumption as shown below:
{
\small
\begin{align*}
    |\rvx| + \max \left\{\sum_l \max\{|\rvs_{\text{fwd},l}|, |\rvs_{\text{bwd},l}|\}, |\rvs_{\text{bwd}}| + |\btau_{\text{opt}}| \right\} + |\rvs_{\text{opt}}|.
\end{align*}}%
In the next, we will go through all the FO/ZO optimizers considered in this work and discuss their memory efficiency one by one.

\subsection{Theoretical Memory Efficiency Analysis of Different Optimizers}
\label{sec: theoretical_analysis_mem}

\textbf{FO-SGD.} Following \cref{alg:opt}, the stored model/optimizer states are defined as
\begin{align*}
    \rvs_{\text{fwd}} &= \oplus_{l=1}^{L} \rva_{l} \\
    \rvs_{\text{bwd}} &=  \oplus_{l=1}^{L} \rvg_{l} \\
    \rvs_{\text{opt}} &= \emptyset,
\end{align*}
where $\rva_l$ represents the total activations being stored for computing the backward gradients $\rvg_l$, and $\oplus$ denotes {the vector concatenation operation}.
Therefore, the total memory consumption is
{\small
\begin{align*}
    &\quad |\rvx| + \max \left\{ \sum_l \max\{|\rvs_{\text{fwd},l}|, |\rvs_{\text{bwd},l}|\}, |\rvs_{\text{bwd}}| \right\}+ |\rvs_{\text{opt}}| \\
    &= |\rvx| + \max \left\{ \sum_l \max\{|\rva_l|, |\rvg_l|\}, |\rvg| \right\} + 0 \\
    &= |\rvx| + \sum_l \max\{|\rva_l|, |\rvg_l|\}
\end{align*}}%

\textbf{FO-SGD FP16.}
We use $\bar \rva$ and $\bar \rvx$ to denote the 16-bit version of $\rva$ and $\rvx$ (half-precision), respectively. The states are defined as:
{\small
\begin{align*}
    \rvs_{\text{fwd}} &= \oplus_{l=1}^{L} \left(\bar\rva_{l} \oplus \bar\rvx_{l}\right) \\
    \rvs_{\text{bwd}} &= \oplus_{l=1}^{L} \rvg_{l} \\
    \rvs_{\text{opt}} &= \emptyset.
\end{align*}}%
Here, besides the activations $\bar\rva_l$ stored in float16 for backward computing, there exists an extra copy of model weight in float16 in mixed precision training. This extra copy of model weight is a temporary memory, $|\tau_{\text{fwd}}|= |\bar\rvx|$, during forward computing. Therefore, the total memory consumption is
{
\small
\begin{align*}
    & \quad |\rvx| + \max \left\{|\rvs_{\text{fwd}}| + |\btau_{\text{fwd}}|, |\rvs_{\text{bwd}}| + |\btau_{\text{bwd}}|, |\rvs_{\text{bwd}}|\right\} + |\rvs_{\text{opt}}| \\
    &= |\rvx| + \max \left\{|\bar \rva| + |\bar\rvx|, \sum_l \max\{|\bar \rva_l|, |\rvg_l|\}\right\}.
\end{align*}}%

\textbf{Vanilla ZO-SGD.}
For the vanilla implementation of ZO-SGD, the forward states include the random vector $\rvz$ and projected gradient $\delta$, i.e.,
\begin{align*}
    \rvs_{\text{fwd}} =  \rvz \oplus \delta
\end{align*}
The projected gradient is computed by differentiating two forward passes, where the random vectors are given by:
\begin{align*}
    \rvz = [\rvz_l \sim \mathcal{N}]_{l=1}^L,
\end{align*}
which have the same dimension as the model parameters $\rvx$.
Then, the backward state is the gradients estimated by $\rvz$ and no optimizer state is involved, i.e.,
\begin{align*}
    \rvs_{\text{bwd}} &= \rvg \\
    \rvs_{\text{opt}} &= \emptyset.
\end{align*}
Therefore, the total memory consumption is:
\begin{align*}
   &\quad |\rvx| + \max \left\{ \sum_l \max\{|\rvs_{\text{fwd},l}|, |\rvs_{\text{bwd},l}|\}, |\rvs_{\text{bwd}}| \right\}+ |\rvs_{\text{opt}}| \\
    &= |\rvx| + \max \{|\rvx| + |\rvg|\} + 0 \\
    &= |\rvx| + |\rvg|.
\end{align*}

\textbf{ZO-SGD w/ memory reduction trick (ZO-SGD).}
In this work, the ZO-SGD method is by default implemented with the random state trick outlined in MeZO~\cite{malladi2024finetuning} to reduce the memory consumption. The key idea of this trick is to save the random seed used to generate the random vectors instead of directly saving the random vectors themselves. By manually set the random seed, the same corresponding random vectors for each layer $\rvz_l$ can be generated \textit{on demand} and used for perturbation without any additional storage incurred:
\begin{align*}
    \texttt{rng} = \text{randomState}(\mathcal{S}),\ \rvz = [\rvz_l \sim \mathcal{N}_{\text{rng}}]_{l=1}^L
\end{align*}
where $\texttt{rng}$ is a pseudo-random variable generator. Therefore, all the states needed are:
\begin{align*}
    \rvs_{\text{fwd}} &= \oplus_{l=1}^{L} \rvx_{l} \\
    \rvs_{\text{bwd}} &= \oplus_{l=1}^{L} \rvg_{l} \\
    \rvs_{\text{opt}} &= \emptyset,
\end{align*}
and this yields the total memory consumption:
{\small
\begin{align*}
   &\quad |\rvx| + \max \left\{ \sum_l \max\{|\rvs_{\text{fwd},l}|, |\rvs_{\text{bwd},l}|\}, |\rvs_{\text{bwd}}| \right\}+ |\rvs_{\text{opt}}| \\
    &= |\rvx| + \max \{\rvg_l\}
\end{align*}}%
For simplicity, we use the name ZO-SGD to denote ZO-SGD w/ memory reduction trick throughout the paper.

\textbf{ZO-SGD-Momentum.}
ZO-SGD with Momentum is similar to ZO-SGD, but it consumes extra memory for momentum storage, which shares the same size as the model parameter. This is considered as an optimizer state in \cref{alg:opt}. Therefore, the states are defined as:
\begin{align*}
    \rvs_{\text{fwd}} & = \oplus_{l=1}^{L} \rvx_{l} \\
    \rvs_{\text{bwd}} & = \oplus_{l=1}^{L} \rvg_{l} \\
    \rvs_{\text{opt}} & = \oplus_{l=1}^{L} \rvx_{l}.
\end{align*}
Similar to ZO-SGD, the total memory consumption is:
{\small
\begin{align*}
    &\quad |\rvx| + \max \left\{ \sum_l \max\{|\rvs_{\text{fwd},l}|, |\rvs_{\text{bwd},l}|\}, |\rvs_{\text{bwd}}| \right\}+ |\rvs_{\text{opt}}| \\
    & = |\rvx| + \max \{\rvg_l\} + |\rvx| \\
    & = 2|\rvx| + \max \{\rvg_l\}.
\end{align*}}%

\textbf{Forward Gradient.}
For FG, the forward states include the random vector $\rvz$ and projected gradient $\delta$.
The projected gradient is computed by the Jacobian-Vector Product (JVP) and forward gradient, \textit{i.e.}, $\frac{\partial^\top f}{\partial \rvx} \rvz$, where the random vector is:
\begin{align*}
    \rvz = [\rvz_l \sim \mathcal{N}]_{l=1}^L.
\end{align*}
In addition to $\rvz$ and $\delta$, JVP itself needs to store intermediate results.
For example, consider a simplified two-layer network $\rvy = f_2(f_1(\rvx))$.
Suppose the output of the first layer is $\rva = f_1(\rvx)$.
The corresponding JVP can be formulated as
\begin{align*}
    \frac{\partial \rvy}{\partial \rvx}^\top \rvv &= \frac{\partial \rvy}{\partial \rva}^\top \left( \frac{\partial \rva}{\partial \rvx}^\top \rvv \right),
\end{align*}
where $\left( \frac{\partial \rva}{\partial \rvx}^\top \rvv \right)$ is computed in the first layer and needs to take the memory of size $|\rva|$.
Now, we can summarize the memory required for the forward state $\rvs_{\text{fwd}}$ as
\begin{align*}
    \rvs_{\text{fwd}} =  \rvz \oplus \rva \oplus \delta.
\end{align*}
Then, the backward state is the gradients estimated by $\rvz$, \textit{i.e.},
\begin{align*}
     \rvs_{\text{bwd}} & = \oplus_{l=1}^L \left(\delta \rvz_l\right) \\
     \rvs_{\text{opt}} & = \emptyset.
\end{align*}
Therefore, the total memory consumption is 
\begin{align*}
    &\quad |\rvx| + \max \left\{ \sum_l \max\{|\rvs_{\text{fwd},l}|, |\rvs_{\text{bwd},l}|\}, |\rvs_{\text{bwd}}| \right\}+ |\rvs_{\text{opt}}| \\
    & = |\rvx| + \max \left\{\rvg + \max \nolimits_l \rva_l, \rvg\right\} + 0 \\
    & = |\rvx| + |\rvg| + \max \nolimits_l \{\rva_l\}.
\end{align*}

Forward Gradient (FG) without State is not supported in PyTorch.
This is because PyTorch requires the $\rvz$ to be pre-computed, thus the $\rvs_{\text{fwd}}$ cannot be reduced to stateless implementation.
Therefore, there is no implementation if we are using the forward gradient provided by PyTorch forward-mode automatic differentiation (forward-mode AD).

Though not possible with the PyTorch built-in Automatic Differentiation tools, FG without State is still possible.
Specifically, for each layer, we keep the order when the $\rvz_l$ is computed, then the $\rvz_l$ can be computed on demand. 
The pursuit of a more memory-efficient FG computation in practice remains an open question, reserved for future work.

\subsection{Other Implementation Details}
\label{sec: implementation_details}

\textbf{Half precision (F16).}
Most modern LLMs are served at 16-bit float precision.
By default, models will be loaded at full precision, i.e., 32-bit. 
To speed up inference and improve memory efficiency, models can be loaded at 16-bit precision, namely \emph{F16}.
We do F16 for ZO methods that do not require differentiation.
Note for Forward Gradient, the F16 cannot be loaded for auto differentiation.
For FO methods, the half-precision is not allowed for the same reason.

\textbf{Mixed precision (FP16).}
For FO methods, mixed precision is a common practice to speed up gradient computation and reduce memory complexity.
We denote the mixed precision (i.e., 16-bit on computing gradients) as FP16.
Note that FP16 will not reduce the memory consumption for storing models or gradients but only affect the gradient computation and intermediate results.

For FO-SGD (FP16), the memory consumption for computing gradient is
\begin{align*}
    \max \{\frac {1}{2}|\rva|+\frac {1}{2}|\rvx|, \sum_l \max\{\frac{1}{2} |\rva_l|, |\rvx_l|\}\},
\end{align*}
where $\frac {1}{2}|\rva|+\frac {1}{2}|\rvx|$ is the memory for model and activations in 16 bit and $\max\{\frac{1}{2} |\rva_l|, |\rvx_l|\}$ is the memory of activation and full-precision gradient.

\textbf{The `foreach' implementation of Adam.}
The PyTorch implementation of Adam will use the so-called `foreach' implementation to speed up the computation. At a high level, the foreach implementation will merge all the layers' weight into one tensor during Adam updates.
Though foreach can speed up computation, it will demand extra memory to store all the weights.

\begin{table}[th]
    \centering
    \caption{Memory consumption comparison for \textbf{full fine-tuning}~(FT) evaluated on OPT-13B model with the MultiRC dataset (400 tokens per example on average). Notations are consistent with Tab.\,\ref{tab:total_mem_com}. The theoretical and empirical values below can be mutually corroborated, where $|\rvx| \approx 48$ GB, $\sum_l \max\{|\rva_l|, |\rvx_l|\} \approx 49$ GB. Note $|\rva|$ are dependent on the sequence length of the input.}
    \label{tab:total_mem_ft_measurement}
    \vspace{5pt}
    \resizebox{\linewidth}{!}{
    \begin{tabular}{l|cr}
    \toprule
    \midrule
    Optimizer & Theoretical Mem. & Empirical Mem. \\
    \midrule
    FO-SGD & $\sum_l \max\{|\rva_l|, |\rvx_l|\} + |\rvx |$ & $97$ GB \\
    FO-Adam w/o fast foreach & $\sum_l \max\{|\rva_l|, |\rvx_l|\} + 3|\rvx|$ & $195$ GB \\
    FO-Adam & $\sum_l \max\{|\rva_l|, |\rvx_l|\} + 4|\rvx|$ & $239$ GB\\
    Forward Grad & $2 |\rvx| + \max_l |\rva_l|$ & $103$ GB \\
    Vanilla ZO-SGD & $2|\rvx|$ & $96$ GB\\
    ZO-SGD & $\max_l |\rvx_l| + |\rvx|$ & $51$ GB\\
    ZO-SGD MMT & $\max_l |\rvx_l| + 2|\rvx|$ & $100$ GB \\
    ZO-Adam & $\max_l |\rvx_l| + 3|\rvx|$ & $151$ GB \\
    \midrule
    FO-SGD (\textbf{FP16})  & $\max \left\{\frac {1}{2}|\rva|+\frac {1}{2}|\rvx|, \sum_l \max\{\frac{1}{2} |\rva_l|, |\rvx_l|\}\right\} + |\rvx |$ & $98$ GB \\
    FO-Adam (\textbf{FP16})   & $\max \left\{\frac {1}{2}|\rva|+\frac {1}{2}|\rvx|, \sum_l \max\{\frac{1}{2} |\rva_l|, |\rvx_l|\}\right\} + 4 |\rvx |$ & $239$ GB \\
    \midrule
    ZO-SGD (\textbf{F16}) & $\max_l \frac{1}{2} |\rvx_l| + \frac{1}{2} |\rvx|$ & $25$ GB \\
    ZO-SGD-MMT (\textbf{F16}) & $\max_l \frac{1}{2} |\rvx_l| + |\rvx|$ & $49$ GB \\
    ZO-Adam (\textbf{F16}) & $\max_l \frac{1}{2} |\rvx_l| + \frac{3}{2}|\rvx|$ & $74$ GB \\
    \midrule
    \bottomrule
    \end{tabular}}
\end{table}

\begin{table}[th]
    \centering
    \caption{Memory consumption comparison for \textbf{LoRA} fine-tuning evaluated on OPT-13B model with the MultiRC dataset (400 tokens per example on average). Other settings are consistent with Tab.\,\ref{tab:total_mem_ft_measurement}. The memory consumption of the LoRA adapter's parameters is assumed to be significantly smaller than $|\rvx|$ and $|\rva|$, and thus is omitted in this table.}
    \label{tab:total_mem_lora_measurement}
    \vspace{5pt}
    \resizebox{\linewidth}{!}{
    \begin{tabular}{l|cr}
    \toprule
    \midrule
    Optimizer & Theoretical Mem. & Empirical Mem. \\
    \midrule
    FO-SGD & $\sum_l \max\{|\rva_l|, |\rvepsilon_l|\} + |\rvx|$ & $69$ GB \\
    FO-Adam w/o fast foreach & $\sum_l \max\{|\rva_l|, |\rvepsilon_l|\} + 2|\rvepsilon| + |\rvx|$ & $69$ GB \\
    FO-Adam & $\sum_l \max\{|\rva_l|, |\rvepsilon_l|\} + 3|\rvepsilon| + |\rvx|$ & $69$ GB\\
    Forward Grad & $ |\rvepsilon| + |\rvx| + \max_l |\rva_l|$ & $55$ GB \\
    Vanilla ZO-SGD & $|\rvx| + |\rvepsilon|$ & $52$ GB\\
    ZO-SGD & $\max_l |\rvepsilon_l| + |\rvx|$ & $52$ GB\\
    ZO-SGD MMT & $\max_l |\rvepsilon_l| + |\rvepsilon| + |\rvx|$ & $52$ GB \\
    ZO-Adam & $\max_l |\rvepsilon_l| + 2|\rvepsilon| + |\rvx|$ & $52$ GB \\
    \midrule
    FO-SGD (\textbf{FP16})  & $\max \left\{\frac {1}{2}|\rva|+\frac {1}{2}|\rvx|, \sum_l \max\{\frac{1}{2} |\rva_l|, |\rvepsilon_l|\}\right\} + |\rvx |$ & $92$ GB \\
    FO-Adam (\textbf{FP16})   & $\max \left\{\frac {1}{2}|\rva|+\frac {1}{2}|\rvx|, \sum_l \max\{\frac{1}{2} |\rva_l|, |\rvepsilon_l|\}\right\} + 3|\rvepsilon| + |\rvx |$ & $93$ GB \\
    \midrule
    ZO-SGD (\textbf{F16}) & $\max_l \frac{1}{2} |\rvepsilon_l| + \frac{1}{2} |\rvx|$ & $25$ GB \\
    ZO-SGD-MMT (\textbf{F16}) & $\max_l \frac{1}{2} |\rvepsilon_l| + \frac{1}{2} |\rvepsilon| + \frac{1}{2} |\rvx|$ & $25$ GB \\
    ZO-Adam (\textbf{F16}) & $\max_l \frac{1}{2} |\rvepsilon_l| + |\rvepsilon| + \frac{1}{2}|\rvx|$ & $25$ GB \\
    \midrule
    \bottomrule
    \end{tabular}}
\end{table}

\subsection{Additional Experiments}

We compare the empirical memory costs of full fine-tuning in \textbf{Tab.\,\ref{tab:total_mem_ft_measurement}} and LoRA fine-tuning in \textbf{Tab.\,\ref{tab:total_mem_lora_measurement}}. Notably, the empirical results are generally aligned with the theoretical analysis. ZO methods are generally much more efficient than their FO counterparts, and the overhead can be further reduced by loading the half-precision model (using F16).

\end{document}